\documentclass{article}

\usepackage{PRIMEarxiv}

\usepackage[utf8]{inputenc}
\usepackage[T1]{fontenc}
\usepackage{lmodern}

\usepackage{amsmath, amssymb, amsfonts}
\usepackage{booktabs}
\usepackage{siunitx}
\usepackage{graphicx}
\graphicspath{{media/}}
\usepackage{tikz}
\usetikzlibrary{positioning, cd}

\usepackage{algorithm}
\usepackage{algpseudocode}

\usepackage{microtype}
\usepackage{nicefrac}
\usepackage{enumitem}
\usepackage{float}
\usepackage{titlesec}
\usepackage{caption}
\usepackage{subcaption}
\usepackage{lipsum}

\usepackage{hyperref}
\hypersetup{
  colorlinks  = true,
  linkcolor   = blue,
  citecolor   = blue,
  urlcolor    = blue
}

\usepackage{fancyhdr}
\title{\textbf{CS-SHRED}: Enhancing \textbf{SHRED} for Robust Recovery of Spatiotemporal Dynamics
\thanks{\textit{\underline{Citation}}: 
\textbf{da Silva R. B., Passos, D.; Oishi, C. M.; Kutz, J.N.  CS-SHRED: Enhancing SHRED for Robust Recovery of Spatiotemporal Dynamics.}} 
}

\author{
  Romulo Brito da Silva  \\
  Universidade Estadual Paulista Júlio de Mesquita Filho, \\ Faculdade de Ciências e Tecnologia de Presidente Prudente\\
  UNESP \\
  Presidente Prudente\\
  \texttt{romulo.silva@peq.coppe.ufrj.br} \\
   \And
  Diego Passos\\
  Departamento de Informática\\
  Instituto Superior de Engenharia de Lisboa, \\
  Instituto Politécnico de Lisboa
  Lisbon, Portugal\\
  \texttt{diego.passos@isel.pt} \\
   \And   
  Cassio Machiaveli Oishi \\
  Universidade Estadual Paulista Júlio de Mesquita Filho,\\Faculdade de Ciências e Tecnologia de Presidente Prudente\\
  UNESP \\
  Presidente Prudente\\
  \texttt{cassio.oishi@unesp.br} \\
  \And
  JN Kutz \\
  Department of Applied Mathematics and Electrical and Computer Engineering, \\University of Washington, Seattle, Washington 98195, USA \\
  \texttt{kutz@uw.edu} \\
}

\begin{document}

\maketitle

\begin{abstract}

We present \textbf{CS-SHRED}, a novel deep learning architecture that integrates Compressed Sensing (CS) into a Shallow Recurrent Decoder (\textbf{SHRED}) to reconstruct spatiotemporal dynamics from incomplete, compressed, or corrupted data. Our approach introduces two key innovations. First, by incorporating CS techniques into the \textbf{SHRED} architecture, our method leverages a batch-based forward framework with $\ell_1$ regularization to robustly recover signals even in scenarios with sparse sensor placements, noisy measurements, and incomplete sensor acquisitions. Second, an adaptive loss function dynamically combines Mean Squared Error (MSE) and Mean Absolute Error (MAE) terms with a piecewise Signal-to-Noise Ratio (SNR) regularization, which suppresses noise and outliers in low-SNR regions while preserving fine-scale features in high-SNR regions.
%
We validate \textbf{CS-SHRED} on challenging problems including viscoelastic fluid flows, maximum specific humidity fields, sea surface temperature distributions, and rotating turbulent flows. Compared to the traditional \textbf{SHRED} approach, \textbf{CS-SHRED} achieves significantly higher reconstruction fidelity---as demonstrated by improved SSIM and PSNR values, lower normalized errors, and enhanced LPIPS scores---thereby providing superior preservation of small-scale structures and increased robustness against noise and outliers.
Our results underscore the advantages of the jointly trained CS and SHRED design architecture which includes an LSTM sequence model for characterizing the temporal evolution with a shallow decoder network (SDN) for modeling the high-dimensional state space.   The SNR-guided adaptive loss function for the spatiotemporal data recovery establishes \textbf{CS-SHRED} as a promising tool for a wide range of applications in environmental, climatic, and scientific data analyses.
\end{abstract}


\section{Introduction}

The \textbf{SHRED} (SHallow REcurrent Decoder) \cite{WILLIAMS2023} framework offers a novel approach to reconstructing spatiotemporal dynamics using multivariate time series from sensor measurements. Recognized for its data-driven nature, \textbf{SHRED} demonstrates robustness even when sensors are mobile~\cite{ebers2024leveraging} or not optimally placed.  Indeed, the \textbf{SHRED} architecture leverages the Taken's embedding theorem and separation of variables in order to produce stable spatio-temporal reconstructions and forecasts from minimal sensor measurements. The \textbf{SHRED} architecture processes multivariate sensor time series data through Long Short-Term Memory (LSTM) recurrent layers. A fully connected, shallow decoder receives the latent representation from the LSTM, aiding in reconstruction and restoration to the high-dimensional state. Across various datasets—including isotropic turbulent flow, sea surface temperature, and atmospheric ozone concentration—\textbf{SHRED} outperforms existing linear and nonlinear methods \cite{WILLIAMS2023}. Remarkably, \textbf{SHRED} achieves accuracy comparable to methods requiring a larger number of optimally placed sensors, even with just three randomly placed sensors for 2D spatio-temporal data. Furthermore, \textbf{SHRED} provides real-time compression, independent of the underlying physics, for modeling engineering and physical systems with low input requirements.  \textbf{SHRED} can be used for reduced order modeling~\cite{tomasetto2025reduced}, model discovery~\cite{gao2025sparse} and digital twin architectures~\cite{riva2024robust}.

Although robust to noisy sensor measurements,  the \textbf{SHRED} model shows limitations in handling spatiotemporal data with gaps in the sensor measurements, corrupt measurements, or insufficient sampling at various time points. These shortcomings become evident when employing a methodology designed to simulate adverse conditions of information loss during sensor data acquisition.  In our approach, \textit{Compressed Sensing} (CS) is employed to mitigate the limitations of practical data-acquisition systems.  Specifically, we perform \textbf{random subsampling} that (i) removes a fixed percentage of spatial samples and (ii) applies this removal only to a randomly selected subset of snapshots. As a result, the remaining measurements come from only a few sensors, sparsely positioned and suboptimally distributed across the domain, producing sparse and irregularly subsampled spatiotemporal series.  This process mimics scenarios in which data are collected incompletely or inadequately due to various real-world factors.
In practical applications, data collection often experiences information loss due to sensor malfunctions, limited coverage, or environmental disturbances. For example, environmental monitoring measurements can be compromised by adverse weather, interference from nearby objects, or sensor technical issues. Similarly, in medical imaging, artifacts or patient movement during scanning may result in incomplete or distorted data.

Thus to address these limitations, we propose integrating the \textit{Compressed Sensing} (CS) framework into \textbf{SHRED}, creating \textbf{CS-SHRED} (Compressed Sensing SHRED). This integration enhances the model's ability to recover missing or incomplete data points while maintaining \textbf{SHRED}'s efficiency in reconstructing spatiotemporal dynamics from a limited number of sensors. The approach introduces a restriction operator that confines the signal to observed (non-null) elements, focusing the recovery problem on known data. The formulation combines both the restriction operator and a sparsifying operator within a sparse minimization framework using convex relaxation, facilitating the accurate recovery of the signal captured by the sparse randomly placed sensors.
Integrating \textbf{SHRED} with the unsupervised learning framework of CS significantly advances the recovery of spatiotemporal dynamics using a limited number of sensors without requiring prior knowledge of the governing laws. Additionally, we develop a novel loss function that incorporates regularization terms related to the signal-to-noise ratio (SNR). This loss function combines traditional L1 and L2 regularization with an SNR-based term, allowing the model to prioritize areas with higher SNR and penalize those with lower SNR. By focusing on more reliable data, the model's overall reconstruction performance is improved, enhancing its resilience to noise and data imperfections.

The ability to handle missing data is particularly relevant in fields such as exploration seismology, where data gaps can compromise interpretation and the identification of oil and gas reservoirs~\cite{hennenfent2008, assis2022compressive}. Similarly, in medical imaging, such as magnetic resonance imaging (MRI), missing information may distort images and impact medical diagnosis~\cite{xia2021}.  To address these challenges, CS has been widely employed as an effective technique for recovering missing information in subsampled or corrupted data. CS leverages the concept that many real-world signals are naturally sparse or can be approximated by a small number of coefficients in a suitable basis. By reconstructing missing data using CS, it is possible to achieve satisfactory results in various applications, considering certain limitations and conditions. This contributes to the accurate analysis and interpretation of data in many contexts.  Despite its advantages, CS also has certain limitations that need to be considered. One of the main challenges is the high computational cost associated with the optimization process in the $ l_1 $ norm, particularly in large datasets. This procedure may require substantial computational resources and prolonged execution time, making it impractical in certain scenarios \cite{ZHANG2020}. Furthermore, the performance of CS can be influenced by the quality and representativeness of the input data. If the data do not adhere to the assumption of sparsity or are not properly preprocessed, CS may yield suboptimal or inadequate results \cite{ZHAO2023}. Another drawback is the sensitivity of CS to certain configuration parameters, such as the size of the dictionary used in the sparse representation of the data. Improper selection of these parameters can lead to poor data reconstruction, compromising the usefulness and reliability of the results \cite{AMINI2020}.

In addition to \textit{Compressed Sensing}, various other techniques have been explored for recovering missing data, each with its own advantages and challenges.  
A common approach is the use of Convolutional Neural Networks (CNN) combined with Long- and Short-Term Memory layers (LSTM) \cite{shi2015convolutional}.  
This combination captures both spatial and temporal features of the data.  
In the realm of missing-data recovery, Generative Adversarial Networks (GANs) have gained prominence for their ability to produce synthetic samples that closely resemble real data; such models improve diversity and quality of imputation, yielding more precise reconstructions \cite{HUANG2023110919}. 
\textit{Transfer Learning} has also been investigated by adapting pre-trained models to the specific task of data retrieval, leveraging knowledge from related datasets to enhance performance and generalization \cite{tan2018survey,CHEN2021126573,HASSAN2017}.  
Additionally, autoencoder-based models learn compact representations that enable effective imputation of missing values \cite{Pereira2020ReviewingAF}. 
Finally, attention mechanisms within neural networks can selectively focus on the most relevant parts of a sequence, leading to more accurate reconstructions \cite{DU2023119619,vaswani2017attention}.

The variety of approaches reflects the complexity and diversity of challenges in recovering missing data, emphasizing the ongoing importance of research and development of robust and efficient techniques to address this problem across a wide range of contexts and applications.

The key contributions of this work include:

\begin{enumerate}
    \item \textbf{Integration of \textit{Compressed Sensing} into the SHRED Framework:}\\[1mm]
    We introduce the \textbf{CS-SHRED} (Compressed Sensing SHallow REcurrent Decoder) architecture, which seamlessly integrates CS techniques into the established SHRED model. This integration enables efficient processing and reconstruction of spatiotemporal data from sparse and incomplete measurements using only a limited number of randomly placed sensors. By leveraging the sparsity of the signal, \textbf{CS-SHRED} accurately recovers missing data points and captures complex system dynamics without requiring prior knowledge of the underlying physical laws.
    
    \item \textbf{Innovative SNR-Based Loss Function for Enhanced Robustness:}\\[1mm]
    We propose a novel loss function that combines traditional $L1$ and $L2$ regularization terms with an adaptive Signal-to-Noise Ratio (SNR) component. This SNR-based regularization prioritizes the reconstruction of high-quality data by penalizing regions with low SNR, thereby enhancing the model’s resilience to noise and data imperfections. Consequently, our approach ensures more reliable and precise reconstructions by focusing on the most informative regions of the dataset.
    
    \item \textbf{Efficient and Scalable Data Recovery through Batch Processing:}\\[1mm]
    We incorporate batch processing into the \textbf{CS-SHRED} framework, enabling the recovery of missing information during the forward pass of the model. This strategy allows training with complete signals and facilitates scalable data reconstruction for large-scale datasets with incomplete or sparsely sampled spatiotemporal measurements. By processing data in batches, \textbf{CS-SHRED} learns comprehensive reconstruction patterns that optimize the overall data recovery process while significantly improving computational efficiency.
    
\end{enumerate}

To evaluate the effectiveness and robustness of the proposed \textbf{CS-SHRED} model, we conducted experiments on four diverse datasets that capture complex spatiotemporal dynamics. Specifically, we considered datasets from the domains of {fluid dynamics and climatology}, including \textit{viscoelastic flow}, \textit{turbulent flow}, \textit{sea surface temperature}, and \textit{maximum specific humidity}.

These datasets were selected because they simulate real-world scenarios where data incompleteness is common and precise reconstruction is critical. Our experimental results consistently show that \textbf{CS-SHRED} significantly outperforms the conventional \textbf{SHRED} approach. In particular, reconstructions produced by \textbf{CS-SHRED} not only preserve the intricate spatiotemporal patterns observed in the \textbf{SST} and \textbf{qmax} datasets but also accurately recover $\text{Tr}(\mathbf{C})$ in the viscoelastic Oldroyd-B model and \texttt{TURB-Rot} data, even under challenging subsampling conditions.

Comparative visualizations further reveal that \textbf{CS-SHRED} captures spatial and temporal variations with superior fidelity, yielding more realistic and detailed reconstructions. These findings underscore the importance of integrating advanced CS techniques with specialized decoding architectures---such as the Shallow Recurrent Decoder---to robustify traditional reconstruction methods.

In Section~\ref{Sec:results}, we provide a detailed quantitative and qualitative evaluation of these results, offering in-depth insights into the advantages conferred by the \textbf{CS-SHRED} model across various applications in environmental and engineering data analysis.

\section{Architecture of Compressed Sensing - SHallow REcurrent Decoder (\textbf{CS-SHRED}) }
\label{sec:headings}

The \textbf{SHRED} model \cite{WILLIAMS2023} is developed to recover high-dimensional spatial and temporal data from a reduced number of sensor measurements. This model operates without requiring additional physical information about the observed phenomenon, preserving its generalization and applicability across various contexts. The \textbf{SHRED} architecture is based on two main components: LSTM layers for temporal feature extraction and fully connected layers for spatial reconstruction.

The model considers the high-dimensional state to be reconstructed as $ \mathbf{x}(t) \in \mathbb{R}^n $ from few sensor measurements represented by $ \mathbf{y}(t) = C \mathbf{x}(t) \in \mathbb{R}^m $ for $ t \in \{t_{\text{current}} - l, t_{\text{current}} - l + 1, \ldots, t_{\text{current}} - 1, t_{\text{current}}\} $. Here, $ l $ is the number of lagged time steps determined empirically, and $ C \in \mathbb{R}^{m \times n}$ consists of rows from the $ n \times n $ identity matrix, assuming that the sensors are sparse ($ m \ll n $).

The LSTM layers receive as input the set of measurements $ \{\mathbf{y}(i)\}_{i=t_\text{current}-l}^{t_\text{current}} $ and update the hidden state $ h_t \in \mathbb{R}^p$ according to the following recursive equations:
\begin{align}
h_t &= \sigma \left(W_o h_{t-1} + W_y \mathbf{y}(t) + b_o \right) \odot \tanh(c_t) \\
c_t &= \sigma \left(W_f h_{t-1} + W_y \mathbf{y}(t) + b_f \right) \odot c_{t-1} + \sigma \left(W_i h_{t-1} + W_y \mathbf{y}(t) + b_i \right) \odot \tanh \left(W_g h_{t-1} + W_y \mathbf{y}(t) + b_g \right)
\end{align}
where $ \sigma $ is the sigmoid function, $ \odot $ denotes the Hadamard product, and the trainable parameters are:
\begin{itemize}
    \item $W_o, W_f, W_i, W_g \in \mathbb{R}^{p \times p}$: weight matrices for the output, forget, input, and cell gates respectively 
    \item $W_y \in \mathbb{R}^{p \times m}$: weight matrix for the input transformation
    \item $b_o, b_f, b_i, b_g \in \mathbb{R}^p$: bias vectors
\end{itemize}
These parameters are collectively denoted as $W_{RN}$ in the final hidden state equation:
\begin{equation}
h_{t_{\text{current}}} = \mathcal{G} \left( \{ \mathbf{y}(i) \}_{i=t_{\text{current}}-l}^{t_{\text{current}}}; W_{RN} \right)
\end{equation}

The fully connected layers transform the LSTM output into high-dimensional predictions through:
\begin{equation}
\mathcal{F}(h; W_{SD}) := R \left(W_b R \left(W_{b-1} \cdots R \left(W_1 h \right) \right) \right)
\end{equation}
where $ R $ is the ReLU activation function and $ W_{SD} = \{W_1, \ldots, W_b\} $ are the trainable weight matrices of the decoder, with $W_1 \in \mathbb{R}^{d_1 \times p}, \ldots, W_b \in \mathbb{R}^{n \times d_{b-1}}$, where $d_i$ are the dimensions of the intermediate layers.

The \textbf{CS-SHRED} model extends the base architecture to address the challenge of reconstructing high-dimensional states from sparse and incomplete measurements. Given incomplete real snapshots $\mathbf{x}_\text{sub}(t)$, the model aims to recover a complete high-dimensional state $\mathbf{x}(t_\text{current}) \in \mathbb{R}^n$, where sensor measurements are represented as $\mathbf{y}_\text{sub}(t_\text{current}) = C \mathbf{x}_\text{sub}(t) \in \mathbb{R}^m$. 

Building upon the work of \cite{WILLIAMS2023}, \textbf{CS-SHRED} innovates by simultaneously handling two distinct types of sparsity: one arising from randomly placed sensor measurements and another from incomplete time series data. This dual-sparsity approach, coupled with the assumption that signals are sparse or compressed in some transform domain, makes the model particularly effective for applications with both sparse and corrupted data. The architecture of \textbf{CS-SHRED} can be expressed as follows, Eq.~\ref{eq:csshred}:

\begin{equation}\label{eq:csshred}
\begin{split}
&\mathcal{H}(\{\mathbf{y}_{\text{sub}}(i)\}_{i=t_{\text{current}}-l}^{t_{\text{current}}}) = \\
&\mathcal{F}\left(\mathcal{G}\left(\underset{\{\xi_i\}_{i=t_{\text{current}}-l}^{t_{\text{current}}}}{\text{argmin}}\|\Theta \{\xi_i\}_{i=t_{\text{current}}-l}^{t_{\text{current}}} - \{\bar{\mathbf{y}}_{\text{sub}}(i)\}_{i=t_{\text{current}}-l}^{t_{\text{current}}}\|_2^2 + \lambda\|\{\xi_i\}_{i=t_{\text{current}}-l}^{t_{\text{current}}}\|_1 ; W_{RN} \right); W_{SD}\right)
\end{split}
\end{equation}
where $ \Theta $ is defined as an operator that performs the recovery of missing information by combining the restriction operator $ R_{op} $ of dimension $ m \times n $ and the adjoint operator $ F_{op}^H $ of the Fourier transform $ F_{op} $ of dimension $ n \times n $:
\begin{equation}\label{eq:theta}
  \Theta := R_{op} \cdot F_{op}^H.
\end{equation}

Here, $ \{\xi_i\}_{i=t_{\text{current}}-l}^{t_{\text{current}}} $ represents the frequency domain representation of the signal to be recovered, with each $ \xi_i \in \mathbb{R}^n $. $ \{\bar{\mathbf{y}}_{\text{sub}}(i)\}_{i=t_{\text{current}}-l}^{t_{\text{current}}} $ is the measurement of the actual signal after the application of the restriction operator, where each $ \bar{\mathbf{y}}^{\text{sub}}_i:=\bar{\mathbf{y}}_{\text{sub}}(i) \in \mathbb{R}^m $. The parameter $ \lambda $ is a scalar that, if different from 0, solves the Basis Pursuit Denoising (BPDN) problem \cite{chen_basis_1994} and controls the trade-off between sparsity and reconstruction fidelity.

As in \cite{WILLIAMS2023}, $ W_{RN} $ represents the weights of the LSTM network, and $ W_{SD} $ refers to the weights of the fully connected neural network. The term $ \{\mathbf{y}_{\text{sub}}(i)\}_{i=t_{\text{current}}-l}^{t_{\text{current}}} $ consists of measurements of the high-dimensional state $ \{\mathbf{x}_{\text{sub}}(i)\}_{i=t_{\text{current}}-l}^{t_{\text{current}}} $, where each $ \mathbf{x}_i:=\mathbf{x}(i) \in \mathbb{R}^n $. $ \mathcal{F} $ is a fully connected neural network with output dimension $ n $, while $ \mathcal{G} $ is an LSTM network. The main and significant difference in terms of architecture lies inside the term $ \mathcal{G} $, which solves the following convex minimization problem to recover the lost/incomplete (or compressed) signal:
\begin{equation}\label{eq:recover}
    \underset{\{\xi_i\}_{i=t_{\text{current}}-l}^{t_{\text{current}}}}{\text{argmin}}\|\Theta \{\xi_i\}_{i=t_{\text{current}}-l}^{t_{\text{current}}} - \{\bar{\mathbf{y}}_{\text{sub}}(i)\}_{i=t_{\text{current}}-l}^{t_{\text{current}}}\|_2^2 + \lambda\|\{\xi_i\}_{i=t_{\text{current}}-l}^{t_{\text{current}}}\|_1.
\end{equation}

The term $ \|\Theta \{\xi_i\}_{i=t_{\text{current}}-l}^{t_{\text{current}}} - \{\bar{\mathbf{y}}_{\text{sub}}(i)\}_{i=t_{\text{current}}-l}^{t_{\text{current}}}\|_2^2 $ quantifies the squared difference between the reconstructed signal obtained by applying the $ \Theta $ operator (which consists of the composition of the adjoint Fourier and restriction operators) to the frequency domain representation $ \{\xi_i\}_{i=t_{\text{current}}-l}^{t_{\text{current}}} $, and the actual measurements of the signal $ \{\bar{\mathbf{y}}_{\text{sub}}(i)\}_{i=t_{\text{current}}-l}^{t_{\text{current}}} $ after applying the restriction operator. This term ensures that the reconstruction of the signal adjusts as closely as possible to the observed data. The term $ \|\{\xi_i\}_{i=t_{\text{current}}-l}^{t_{\text{current}}}\|_1 $ promotes sparsity in the representation of the signal in the frequency domain $ \{\xi_i\}_{i=t_{\text{current}}-l}^{t_{\text{current}}} $, encouraging the presence of few non-zero coefficients. 

To tackle this convex minimization issue, the SPGL1 solver (Spectral Projected Gradient for L1 minimization) is used, \cite{van_den_berg_probing_2009}. After solving the convex minimization problem, the recovered signal in the frequency domain, represented as $ \{\xi_i\}_{i=t_{\text{current}}-l}^{t_{\text{current}}} $, is then transformed back to the time domain using the adjoint operator of the Fourier transform. This yields the recovered signal in the time domain, which can be compared to the actual observed data. This transformation is performed by the operation $ \{\mathbf{y}^{\star}(i)\}_{i=t_{\text{current}}-l}^{t_{\text{current}}} = F_{op}^H \cdot \{\xi_i\}_{i=t_{\text{current}}-l}^{t_{\text{current}}} $. The objective is to minimize the $ \ell_2 $ norm of the difference between the recovered signals and the real measurements $ \{\bar{\mathbf{y}}_{\text{sub}}(i)\}_{i=t_{\text{current}}-l}^{t_{\text{current}}} $, while also adding an $ \ell_1 $ regularization term to promote sparsity. All computational formulations are done with the assistance of PyLops \cite{ravasi2020pylops}, NumPy, and PyTorch libraries.

The results of the reconstruction are passed to the fully connected layers to refine the latent representations. These layers transform the representations learned by the LSTM into high-dimensional predictions.

\subsection{Loss Function and Optimization}
The ultimate goal of the network is to minimize the reconstruction loss given by
\begin{equation}\label{eq:reconst_error}
\mathcal{H} \in \underset{\tilde{\mathcal{H}} \in \mathcal{H}}{\text{argmin}}\quad\mathcal{L}(\mathbf{x}(t_{\text{current}}),\tilde{\mathcal{H}}(\{\mathbf{y}^{\star}(t)\}_{t=t_{\text{current}}-l}^{t_{\text{current}}}))
\end{equation}
where the set of training states is $\{\mathbf{x}(t)\}_{t=1}^T$ and their corresponding recovered measurements are $\{\mathbf{y}^{\star}(t)\}_{t=1}^T$ from the minimization problem. Here, $\mathcal{H}$ represents the reconstruction hypothesis obtained by the \textbf{CS-SHRED} model, and the loss function $\mathcal{L}$ in Eq.~\ref{eq:reconst_error} is defined by
\begin{equation}\label{eq:loss}
  \mathcal{L}=
\begin{cases}
    \lambda_{\text{snr}} \cdot \text{snr}^{-1} + \lambda_{\text{L2}} \cdot \mathcal{L}_{\text{MSE}} + \lambda_{\text{L1}} \cdot \mathcal{L}_{\text{MAE}} + \mathcal{R}_{\ell_{2}}, & \text{snr} > 0 \\
    -\lambda_{\text{snr}} \cdot \text{snr} + \lambda_{\text{L2}} \cdot \mathcal{L}_{\text{MSE}} + \lambda_{\text{L1}} \cdot \mathcal{L}_{\text{MAE}} + \mathcal{R}_{\ell_{2}}, & \text{snr} \leq 0
\end{cases}  
\end{equation}
where $\lambda_{\text{L2}}, \lambda_{\text{L1}}, \lambda_{\text{snr}} \in \mathbb{R}^+$ are hyperparameters related to the mean squared error (MSE) loss, mean absolute error (MAE) loss, and SNR loss, respectively. The Signal-to-Noise Ratio (SNR) is computed as 
\begin{equation}
\text{SNR} = 10 \log_{10}\left(\frac{\text{signal power}}{\text{noise power} + \epsilon}\right)
\end{equation}
where $\text{signal power} = \frac{1}{T}\sum_{t=1}^T \left\|\mathcal{H}(\{\mathbf{y}^{\star}(t)\}_{t=t_{\text{current}}-l}^{t_{\text{current}}})\right\|_{2}^2$, $\text{noise power} = \frac{1}{T}\sum_{t=1}^T \left\|\mathbf{x}(t) - \mathcal{H}(\{\mathbf{y}^{\star}(t)\}_{t=t_{\text{current}}-l}^{t_{\text{current}}})\right\|_{2}^2$, and $\epsilon > 0$ is a smoothing term to prevent division by zero.

The first term of the loss function,
$$
\lambda_{\text{snr}} \cdot \text{SNR}^{-1},
$$
weights the signal-to-noise ratio between the obtained reconstruction and the original training state. This term aims to ensure that the reconstruction maintains the quality of the original signal, penalizing low-quality reconstructions relative to noise.

The second term,
$$
\mathcal{L}_{\text{MSE}} = \lambda_{\text{L2}} \cdot \sum_{t=1}^T \left\|\mathcal{H}(\{\mathbf{y}^{\star}(t)\}_{t=t_{\text{current}}-l}^{t_{\text{current}}}) - \mathbf{x}(t)\right\|_{2}^2,
$$
represents the mean squared error between the reconstruction and the training states. It seeks to minimize the overall differences between the reconstruction and the training data.

The third term,
$$
\mathcal{L}_{\text{MAE}} = \lambda_{\text{L1}} \cdot \sum_{t=1}^T \left\|\mathcal{H}(\{\mathbf{y}^{\star}(t)\}_{t=t_{\text{current}}-l}^{t_{\text{current}}})\right\|_{1},
$$
refers to the mean absolute error of the reconstruction. This term promotes sparsity in the reconstruction, encouraging more parsimonious solutions.

Finally, the regularization term $\mathcal{R}_{\ell_{2}}$ aims to prevent overfitting and ensure solution stability. The optimization is performed using the Adam optimizer with weight decay \cite{loshchilov2017decoupled}.

The inclusion of both MSE and SNR terms in the loss function serves distinct purposes. The MSE term focuses on minimizing the average squared differences between the estimated and true values, ensuring that the reconstructed signal closely matches the true signal in terms of magnitude. On the other hand, the SNR term measures the quality of the signal relative to the background noise. It ensures that the reconstructed signal maintains high quality relative to noise, which is particularly important in scenarios where the signal power varies significantly. This dual approach helps achieve both accuracy and robustness in signal reconstruction.

Similar to \cite{WILLIAMS2023}, the reconstruction error for the test set is defined as
\begin{equation}\label{eq:norm2_error}
   \text{Error} = \frac{1}{n_t} \sum_{t=1}^{n_t} \frac{\left\| \mathcal{H}(\{\mathbf{y}^{\star}(t)\}_{t=t_{\text{current}}-l}^{t}) - \mathbf{x}(t) \right\|_{2}^{2}}{\left\| \mathbf{x}(t) \right\|_{2}^{2}}
\end{equation}
where $n_t$ is the number of test samples.


Intrinsically, the \textbf{CS-SHRED} algorithm relies on multivariate time series from sensor measurements to estimate the state, where each dataset is truncated to reconstruct only the last $n_t - l$ temporal snapshots, where $n_t$ is the initial number of samples and $l$ is the sequence length used for the time series. This length can be seen as a hyperparameter that can be adjusted according to the temporal characteristics of the data.

\begin{figure}[htbp]
    \centering
    \includegraphics[width=1.1\linewidth]{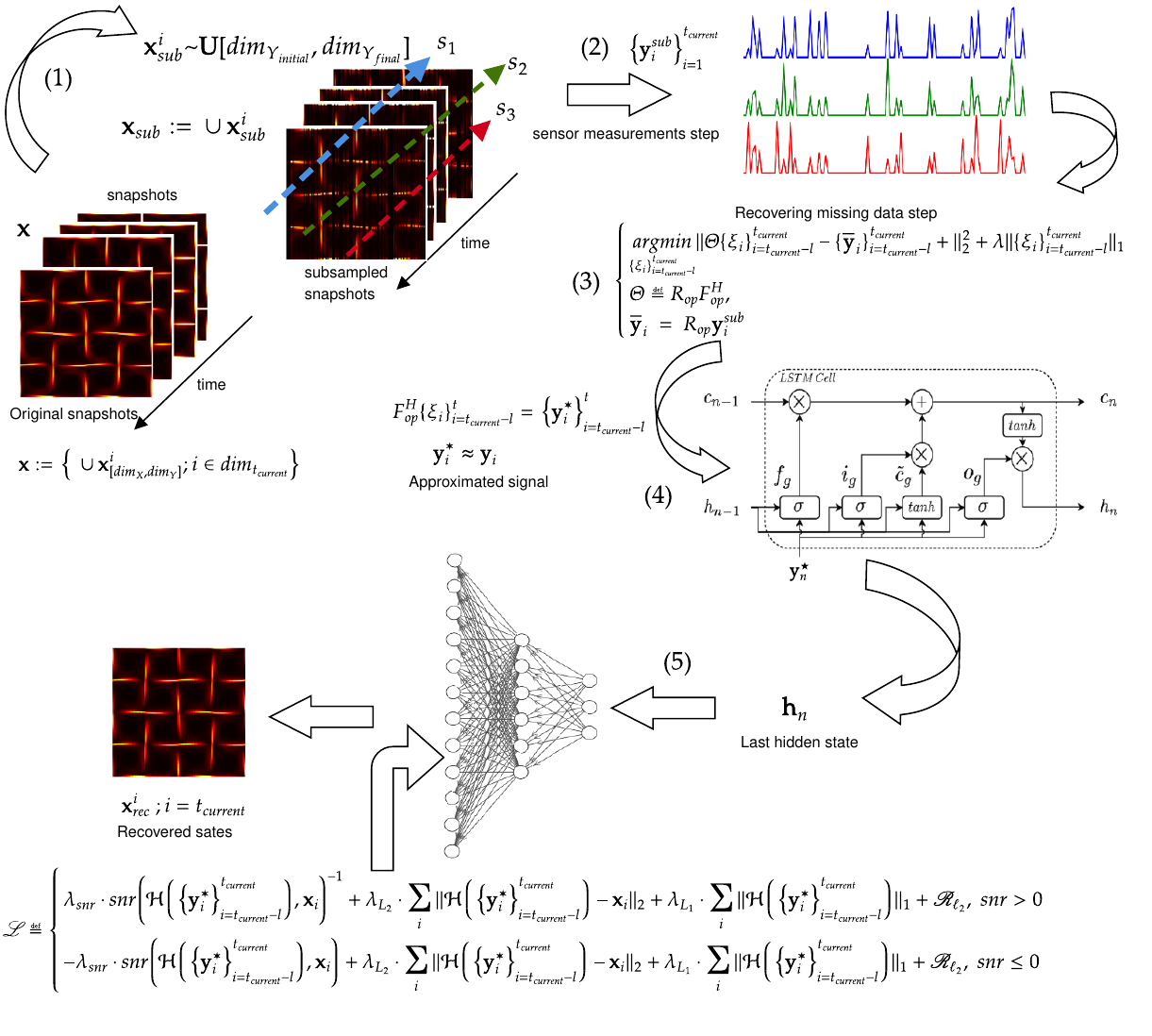}
    \caption{Overview of the \textbf{CS-SHRED} pipeline: (1) The original dynamics $\mathbf{x}$ are uniformly sub-sampled in one spatial dimension for each snapshot. (2) A few sub-sampled sensors $(s_1, s_2, s_3)=\mathbf{y}_i^{\text{sub}}$ are randomly positioned to capture the multivariate time series $\{\mathbf{y}^\text{sub}_i\}_{i=t_\text{current}-l}^{t_\text{current}}$ with a predefined percentage of missing information. (3) The missing data in the multivariate time series collected by the sensors are recovered for each training batch by solving a convex minimization problem. (4) The estimated multivariate time series $\{\mathbf{y}^\star_i\}_{i=t_\text{current}-l}^{t_\text{current}}$ serves as input to the LSTM network, which outputs the final hidden state $\mathbf{h}_n$. (5) This hidden state is then input into the fully connected shallow decoding network, which maps the hidden state to the high-dimensional space generated by the input sequences captured by the sensors.}
    \label{fig:pipe_CS-SHRED}
\end{figure}

The reconstruction pipeline (Fig.~\ref{fig:pipe_CS-SHRED}) of the spatiotemporal dynamics using \textbf{CS-SHRED} consists of five main steps:

1. Original Data: The process starts with the original data $\mathbf{x} := \{\mathbf{x}^t(\text{dim}_x,\text{dim}_y); t \in \text{dim}_T\}$.

2. Subsampling: The subsampling strategy Eq.~\ref{eq:subsample} models practical scenarios in sensor networks where data acquisition is limited by systematic sensor failures and spatial coverage constraints. A subset of spatial columns (dimension $\text{dim}_y$) is randomly selected to be subsampled in specific time snapshots, resulting in subsampled snapshots $\mathbf{x}_\text{sub}$:
\begin{equation}\label{eq:subsample} 
\mathbf{x}_\text{sub}(x, y, t) = 
\begin{cases}
0 & \text{if } y \in \mathcal{Y}_\text{sub} \text{ and } t \in \mathcal{T}_\text{sub}, \\
\mathbf{x}(x, y, t) & \text{otherwise},
\end{cases}
\end{equation}

where $\mathcal{Y}_\text{sub} \subset \{1,\ldots,n_y\}$ denotes the subset of spatial locations where sensor malfunctions occurred, with the total number of affected locations given by $|\mathcal{Y}_\text{sub}| = n_{\text{cols}}$. Similarly, $\mathcal{T}_\text{sub} \subset \{1,\ldots,n_t\}$ represents the time instances corresponding to periods of systematic failures or planned deactivations, with $|\mathcal{T}_\text{sub}| = n_{\text{snap}}$, and it always includes the final snapshot (for convenience). This formulation emulates real-world conditions in which sensor data may be intermittently unavailable due to factors such as maintenance cycles, power management protocols in wireless sensor networks, communication disruptions, or constraints in data transmission.

3. Sensor Measurements: The process yields measurements $\{\mathbf{y}_\text{sub}(i)\}_{i=t_\text{current}-l}^{t_\text{current}}$ from the sensors $(s_1, s_2, s_3,\cdots,s_m)$, where sensor placement is often constrained by physical installation limitations, implementation costs, and site accessibility. For simplicity, Fig.~\ref{fig:pipe_CS-SHRED} illustrates only three sensors $(s_1, s_2, s_3)$.

4. Data Recovery: Missing data are recovered through a convex minimization problem that leverages the spatiotemporal structure of the data:
$$
\mathbf{y}^\star(i) = \mathbf{y}^\star_i = \arg \min_{\{\xi_i\}_{i=t_\text{current}-l}^{t_\text{current}}} \left\|\Theta \left( \{\xi_i\}_{i=t_\text{current}-l}^{t_\text{current}} - \{\bar{\mathbf{y}}_\text{sub}(i)\}_{i=t_\text{current}-l}^{t_\text{current}} \right) \right\|_2^2 + \lambda \left\| \{\xi_i\}_{i=t_\text{current}-l}^{t_\text{current}} \right\|_1
$$

5. LSTM and Decoder: The recovered time series $\{\mathbf{y}^\star(i)\}_{i=t_\text{current}-l}^{t_\text{current}}$ is processed by the LSTM network, producing the hidden state $\mathbf{h}_n$. This state is then decoded by the shallow network to reconstruct the full spatiotemporal dynamics $\mathbf{x}_\text{rec}^t$.

This systematic approach to data reconstruction preserves the essential spatiotemporal structure of the measurements while accounting for practical limitations in real-world sensor networks. The methodology is particularly relevant for applications such as oceanic monitoring, meteorological sensor networks, industrial monitoring, and urban sensing, where similar patterns of data unavailability are frequently observed.

\subsubsection{Implementation Details of \textbf{CS-SHRED}}\label{sec:implementation}

The signal recovery process, detailed in Algorithm \ref{algo:recover_signal}, consists of two main functions: \texttt{recover\_signal\_per\_column} and \texttt{recover\_signal}. The \texttt{recover\_signal\_per\_column} function processes an input tensor $\{\mathbf{y}(t)\}_{t=t_{\text{current}}-l}^{t_{\text{current}}}$ by iterating through its spatial dimensions ($\text{dim}_x$ and $\text{dim}_y$). For each spatial location $(i,j)$, it extracts the corresponding time series data and applies the \texttt{recover\_signal} function to reconstruct any missing values. The recovered signals are collected and returned as a complete set of reconstructed time series $\{\mathbf{y}^{\star}(t)\}_{t=t_{\text{current}}-l}^{t_{\text{current}}}$.

The core signal reconstruction is performed by the \texttt{recover\_signal} function, which takes as input a time series $\mathbf{y}_{\text{sub}}(t)$ and a penalty parameter $\lambda$ that controls the sparsity of the solution. The function first identifies the indices ($iava$ - \textit{indices available}) where the signal has non-zero values, representing the available observations. Using these indices, it constructs a restriction operator $R_{\text{op}}$ defined as:

\begin{equation}\label{eq:rop}
    R_{\text{op}} = R \cdot \mathbf{y}_{\text{sub}}(t)
\end{equation}

where $R$ is a sparse matrix with elements:

$$
R_{ij} = \begin{cases}
1, & \text{if } j = \text{iava}[i] \\
0, & \text{otherwise}
\end{cases}
$$

This restriction operator selects only the observed (non-zero) elements of the signal. The function then defines a Fourier operator $F_{\text{op}}$ to transform the signal into the frequency domain, where it is expected to have a sparse representation. The composite operator $\Theta = R_{\text{op}} \times F_{\text{op}}^H$ combines the restriction and Hermitian conjugate of the Fourier operator.

The signal recovery problem is formulated as a basis pursuit denoising optimization:

$$
\mathbf{y}^{\star}(t) = \arg\min_{\xi} \|\Theta \cdot \xi - \bar{\mathbf{y}}_{\text{sub}}(t)\|_2^2 + \lambda \cdot \|\xi\|_1
$$

where $\bar{\mathbf{y}}_{\text{sub}}(t)$ represents the observed values ($R_{\text{op}} \times \mathbf{y}_{\text{sub}}(t)$). This optimization problem is solved using the SPGL1 algorithm, which balances the fidelity to observed data (first term) against the sparsity of the solution in the Fourier domain (second term). The $\lambda$ parameter controls this trade-off. This approach effectively recovers missing values by exploiting the signal's sparse structure in the frequency domain, even when significant portions of the data are missing.

\begin{algorithm}[t]
  \caption{Signal Recovery}
  \begin{algorithmic}[1]
    \Function{recover\_signal\_per\_column}{$\mathbf{y}_{\text{sub}}(t), \lambda$}
      \State $recovered\_signals \gets \{\}$
      \For{$i \in \{1, 2, \dots, \text{dim}_x\}$}
        \For{$j \in \{1, 2, \dots, \text{dim}_y\}$} \Comment{Iterate through spatial dimensions}
          \State $column\_data \gets \mathbf{y}_{\text{sub}}[:, i, j]$ \Comment{Extract the column data}
          \State $recovered\_signal \gets$ \Call{recover\_signal}{$column\_data, \lambda$} \Comment{Recover the signal for the column}
          \State Add $recovered\_signal$ to $recovered\_signals$
        \EndFor
      \EndFor
      \State \Return $recovered\_signals$
    \EndFunction
    \Statex
    \Function{recover\_signal}{$\mathbf{y}_{\text{sub}}(t), \lambda$}
      \State Let $iava$ be the set of indices where $\mathbf{y}_{\text{sub}}(t) > 0$
      \State Construct the restriction operator: $R_{\text{op}}$ with size $|\mathbf{y}_{\text{sub}}(t)|$ and indices $iava$
      \State Compute $\bar{\mathbf{y}}_{\text{sub}}(t) \gets R_{\text{op}} \times \mathbf{y}_{\text{sub}}(t)$
      \State Define the Fourier operator: $F_{\text{op}} = \text{FFT}(|\mathbf{y}_{\text{sub}}(t)|)$
      \State Define the operator $\Theta$: $\Theta = R_{\text{op}} \times F_{\text{op}}^H$
      \State Solve the basis pursuit denoising problem using SPGL1 to obtain $\mathbf{y}^{\star}(t)$, where:
      \State $\mathbf{y}^{\star}(t) = \arg\min_{\xi} \|\Theta \cdot \xi - \bar{\mathbf{y}}_{\text{sub}}(t)\|_2^2 + \lambda \cdot \|\xi\|_1$
      \State \Return $\mathbf{y}^{\star}(t)$
    \EndFunction
  \end{algorithmic}
  \label{algo:recover_signal}
\end{algorithm}

The \textbf{CS-SHRED} model, depicted in Algorithm \ref{algo:csshred}, enhances the original \textbf{SHRED} model by incorporating a compressive sensing-based strategy to effectively handle the recovery of missing data in spatiotemporal signals. This innovative approach allows the model to manage missing data recovery within each training batch, thereby improving its robustness and generalization capabilities.

The recovery process in the \textbf{CS-SHRED} model begins with the \texttt{recover\_signal\_per\_column} function, which is responsible for reconstructing the signal for each column in the input tensor $\{\mathbf{y}_{\text{sub}}(t)\}_{t=t_{\text{current}}-l}^{t_{\text{current}}}$. This function iterates over each column, extracts the relevant data, and utilizes the \texttt{recover\_signal} function to fill in the missing values, providing an approximation of the signal according to the principles of CS. After the signals are recovered, they are combined into a unified representation $\{\mathbf{y}^{\star}(t)\}_{t=t_{\text{current}}-l}^{t_{\text{current}}}$ and fed into the model's LSTM layers and subsequently the dense layers. This integration allows the model to learn (during the training phase) to reconstruct missing data, enhancing its ability to generalize to unseen data.
\begin{algorithm}[t]
  \caption{\textbf{CS-SHRED} Model}
  \begin{algorithmic}[1]
    \Function{forward}{$\{\mathbf{y}_{\text{sub}}(t)\}_{t=t_{\text{current}}-l}^{t_{\text{current}}}$}
      \State $recovered\_signals\_per\_column \gets$ \Call{recover\_signal\_per\_column}{$\{\mathbf{y}_{\text{sub}}(t)\}_{t=t_{\text{current}}-l}^{t_{\text{current}}}$}
      \State $combined\_recovered\_signal \gets \text{combine\_signals}(recovered\_signals\_per\_column)$
      \State $h_0, c_0 \gets \text{initialize\_hidden\_states}()$
      \State $h_{\text{out}} \gets \text{LSTM}(combined\_recovered\_signal, (h_0, c_0))$
      \State $output \gets \text{Linear\_Layers}(h_{\text{out}})$
      \State \Return $output$
    \EndFunction
  \end{algorithmic}
  \label{algo:csshred}
\end{algorithm}

The training process for the \textbf{CS-SHRED} model is implemented in the \texttt{fit\_csshred\_model} function (Algorithm~\ref{alg:train_csshred}), which accepts the model $\mathcal{H}$, training dataset $\mathcal{D}_{train}$, validation dataset $\mathcal{D}_{val}$, and hyperparameters including batch size, number of epochs $N_{epochs}$, learning rate $\eta$, and regularization coefficients ($\lambda_{\text{L2}}$, $\lambda_{\text{L1}}$, $\lambda_{\text{SNR}}$). The model employs an optimizer $\mathcal{O}$ (Adam) with weight decay $w$ and a learning rate scheduler $\mathcal{S}$ that reduces $\eta$ when the validation loss plateaus.

During each training epoch, the model processes data in batches $(\{\mathbf{y}^{\star}(t)\}_{t=t_{\text{current}}-l}^{t_{\text{current}}}, \mathbf{x}(t_{\text{current}})) \in \mathcal{D}_{train}$ and computes three main loss components: Mean Squared Error $\mathcal{L}_{\text{MSE}}$ between predictions $\mathcal{H}(\{\mathbf{y}^{\star}(t)\}_{t=t_{\text{current}}-l}^{t_{\text{current}}})$ and targets $\mathbf{x}(t_{\text{current}})$, L1 loss $\mathcal{L}_{\text{L1}}$ to promote sparsity, and an SNR-based loss that evaluates signal quality. The total loss $\mathcal{L}$ combines these components with L2 regularization $\mathcal{R}_{\ell_2}$:

$$
\mathcal{L} = \begin{cases}
\min\left(\frac{\lambda_{\text{SNR}}}{\text{SNR} + \epsilon}, 100\right) + \lambda_{\text{L2}} \cdot \mathcal{L}_{\text{MSE}} + \lambda_{\text{L1}} \cdot \mathcal{L}_{\text{L1}} + w \cdot \mathcal{R}_{\ell_2} & \text{if } \text{SNR} > 0 \\
-\lambda_{\text{SNR}} \cdot \text{SNR} + \lambda_{\text{L2}} \cdot \mathcal{L}_{\text{MSE}} + \lambda_{\text{L1}} \cdot \mathcal{L}_{\text{L1}} + w \cdot \mathcal{R}_{\ell_2} & \text{otherwise}
\end{cases}
$$

where $w$ is the weight decay coefficient and $\epsilon$ is a small constant for numerical stability. This formulation encourages the model to maximize $\text{SNR}$ while maintaining reconstruction accuracy and sparsity.

The model's performance is evaluated on $\mathcal{D}_{val}$ at specified intervals using the same loss function. An early stopping mechanism monitors the validation loss and terminates training if no improvement is observed after a predefined number of epochs (patience), reverting to the best parameters $\boldsymbol{\theta}_{\text{best}}$.

The methodology employs CS-based subsampling through the \texttt{subsample} function (Eq.~\ref{eq:subsample}) to emulate scenarios where sensor measurements have missing information due to adverse environmental conditions. This preprocessing step systematically creates data gaps, simulating real-world situations where sensor readings might be compromised or lost. The CS component then works to recover the missing information from these incomplete observations by leveraging sparsity principles.

The neural network training process is guided by the specialized loss function $\mathcal{L}$ that focuses on enhancing the quality of the spatio-temporal dynamics reconstruction through the three complementary penalty terms. The synergy between the CS-based signal recovery and the multi-term loss function optimization enables the model to effectively handle scenarios with missing data while achieving high-quality reconstruction through noise treatment and outlier removal. This dual approach ensures both the recovery of missing information and the enhancement of signal quality in the reconstructed data.

\begin{algorithm}[t]
  \caption{Training \textbf{CS-SHRED} Model}
  \begin{algorithmic}[1]
    \Function{fit\_csshred\_model}{$\mathcal{H}, \mathcal{D}_{train}, \mathcal{D}_{val}, \text{hyperparameters}$}
      \State Initialize optimizer $\mathcal{O}$, learning rate scheduler $\mathcal{S}$
      \State Initialize loss functions: $\mathcal{L}_{\text{MSE}}, \mathcal{L}_{\text{L1}}$
      \State $\boldsymbol{\theta}_{\text{best}} \gets \mathcal{H}.\text{parameters}$
      
      \For{$epoch = 1$ to $N_{epochs}$}
        \For{batch in $\mathcal{D}_{train}$}
          \State $\hat{\mathbf{x}}(t_{\text{current}}) \gets \mathcal{H}(\{\mathbf{y}^{\star}(t)\}_{t=t_{\text{current}}-l}^{t_{\text{current}}})$
          \State $\mathcal{L}_{\text{MSE}} \gets \|\hat{\mathbf{x}}(t_{\text{current}}) - \mathbf{x}(t_{\text{current}})\|_2^2$
          \State $\mathcal{L}_{\text{L1}} \gets \|\hat{\mathbf{x}}(t_{\text{current}})\|_1$
          \State $\text{SNR} \gets 10\log_{10}\left(\frac{\|\mathbf{x}(t_{\text{current}})\|_2^2}{\|\hat{\mathbf{x}}(t_{\text{current}}) - \mathbf{x}(t_{\text{current}})\|_2^2}\right)$
          \State $\mathcal{R}_{\ell_2} \gets \sum_{\theta \in \boldsymbol{\theta}} \|\theta\|_2^2$
          
          \If{$\text{SNR} > 0$}
            \State $\mathcal{L} \gets \min\left(\frac{1}{\text{SNR} + \epsilon}, 100\right)\lambda_{\text{SNR}} + \lambda_{\text{L2}}\mathcal{L}_{\text{MSE}} + \lambda_{\text{L1}}\mathcal{L}_{\text{L1}} + w\mathcal{R}_{\ell_2}$
          \Else
            \State $\mathcal{L} \gets -\text{SNR}\lambda_{\text{SNR}} + \lambda_{\text{L2}}\mathcal{L}_{\text{MSE}} + \lambda_{\text{L1}}\mathcal{L}_{\text{L1}} + w\mathcal{R}_{\ell_2}$
          \EndIf
          
          \State Update $\boldsymbol{\theta}$ using $\nabla_{\boldsymbol{\theta}}\mathcal{L}$
        \EndFor
        
        \If{validation step}
          \State Evaluate on $\mathcal{D}_{val}$
          \State Update $\mathcal{S}$ and early stopping
        \EndIf
      \EndFor
      
      \State \Return $\mathcal{H}(\boldsymbol{\theta}_{\text{best}})$
    \EndFunction
  \end{algorithmic}
  \label{alg:train_csshred}
\end{algorithm}

\section{Results}\label{Sec:results}

In the subsequent sections, we address the spatiotemporal reconstruction of non-Newtonian viscoelastic fluids modeled by the Oldroyd-B equation, as well as the recovery of the spatiotemporal dynamics of atmospheric data such as specific humidity \textbf{qmax}, Sea Surface Temperature (\textbf{SST}) and \texttt{TURB-Rot} database.

For viscoelastic fluids, our focus is on reconstructing the trace of the conformation tensor, $ \mathrm{tr}(\mathbf{C}) $, which is fundamental for understanding the fluid's viscoelastic properties and its behavior under different flow conditions. Accurate reconstruction of this tensor trace allows for a deeper comprehension of internal stresses and deformations within the fluid, which is essential for reliable predictions in both industrial and scientific simulations.

Regarding the atmospheric datasets, analyzing the spatial and temporal distribution of specific humidity and sea surface temperature is essential for understanding climate patterns and atmospheric dynamics on a global scale. However, obtaining complete and continuous datasets is often challenging due to limitations in sensor availability and variable environmental conditions. To mitigate these constraints, subsampling techniques are employed, reducing data complexity while preserving the essential characteristics of the systems under study.

In addition to these applications, we have applied our reconstruction methods to a rotating turbulent flow scenario using data from the \texttt{TURB-Rot} database \cite{Biferale2020TURBRotAL}. The rotating turbulent flow was simulated on a $256^3$ grid in a triply periodic domain and subsequently subsampled to emulate practical measurement constraints. Comparative analyses between \textbf{CS-SHRED} and \textbf{SHRED} revealed that while \textbf{CS-SHRED} excels at capturing fine spatial details—evidenced by higher SSIM and PSNR and lower LPIPS values for key snapshots—the \textbf{SHRED} model demonstrates a more consistent performance over the entire temporal sequence. This indicates that \textbf{CS-SHRED} is particularly adept at reconstructing detailed, high-fidelity snapshots even under severe data reduction, whereas \textbf{SHRED} maintains stable reconstruction quality over time.

We evaluate the effectiveness of the \textbf{CS-SHRED} method in comparison to \textbf{SHRED} for reconstructing these datasets from subsampled data. Our results demonstrate that \textbf{CS-SHRED} outperforms \textbf{SHRED} in terms of fidelity and accuracy, particularly in preserving structural details and dynamic behaviors in both the viscoelastic fluid model and the atmospheric datasets. The superior performance of \textbf{CS-SHRED} highlights the importance of advanced reconstruction methods to ensure data integrity in scenarios of data loss or limited acquisition. This makes it a valuable tool for studies requiring high precision in the analysis of complex environmental and fluid dynamics data, and its applicability can also be further explored in other domains.

\subsection{Data Preparation}
 The following dataset are considered in the current work:

\begin{itemize}

    \item \textbf{Viscoelastic Flow:}\\
    We employ numerical simulation data from the Oldroyd-B constitutive model \cite{OISHI2023}, which describes the dynamics of non-Newtonian viscoelastic fluids. This dataset focuses on the trace of the conformation tensor, $\text{Tr}(\mathbf{C})$, a critical indicator of the fluid's elastic stress state. Given its complex nonlinear dynamics and multiple spatial and temporal scales, accurately reconstructing $\text{Tr}(\mathbf{C})$ from sparse and incomplete sensor measurements provides a rigorous test of our model's capability to capture both elastic and viscous features.
    
    \item \textbf{Turbulent Flow:}\\
    The rotating turbulent flow dataset from the \texttt{TURB-Rot} database \cite{Biferale2020TURBRotAL} represents a particularly challenging scenario. Simulated on a $256^3$ grid within a triply periodic domain, the dataset encompasses a wide range of turbulent scales. By applying controlled subsampling---removing 30\% of spatial columns in 30\% of temporal snapshots---this dataset emulates realistic measurement constraints. Our results demonstrate that \textbf{CS-SHRED} is highly effective in reconstructing fine spatial details and dynamic behaviors, outperforming the conventional \textbf{SHRED} model, particularly in preserving temporal consistency and spatial fidelity.
    
    \item \textbf{Sea Surface Temperature (\textbf{SST}):}\\
    The \textbf{SST} dataset, available at \url{https://psl.noaa.gov/thredds/catalog/Datasets/noaa.oisst.v2/catalog.html?dataset=Datasets/noaa.oisst.v2/sst.wkmean.1990-present.nc}, comprises measurements of the ocean's surface temperature---critical for understanding climate patterns, ocean currents, and weather forecasting. Due to frequent gaps caused by cloud cover and satellite limitations, \textbf{SST} provides an ideal testbed for our model's ability to reconstruct incomplete and noisy data.
    
    \item \textbf{Maximum Specific Humidity (\textbf{qmax}):}\\
    Accessible at \url{https://psl.noaa.gov/thredds/catalog/Datasets/20thC_ReanV3/Derived/8XDailies/2mMO/catalog.html?dataset=Datasets/20thC_ReanV3/Derived/8XDailies/2mMO/qmax.2m.8Xday.ltm.nc}, the \textbf{qmax} dataset contains measurements of the maximum specific humidity, a key variable for analyzing moisture distribution and atmospheric processes. Its inherent incompleteness and irregular sampling challenge our model to accurately reconstruct the underlying spatiotemporal patterns.

\end{itemize}

The preparation of datasets for training, validation, and testing was conducted through a series of preprocessing steps to ensure data quality and suitability for both models. The raw spatiotemporal data, including the trace of the conformation tensor $\mathrm{tr}(\mathbf{C})$ for viscoelastic fluids and atmospheric variables such as specific humidity \textbf{qmax}, Sea Surface Temperature and \texttt{TURB-Rot}, were transposed to align the spatial and temporal dimensions appropriately.

The initial dataset $\mathbf{x} \in \mathbb{R}^{n_t \times n_x \times n_y}$ undergoes a systematic subsampling process. Initially, the data is transposed to the format $(n_x, n_y, n_t)$ to facilitate the application of the subsampling mask. The process involves two key components: the selection of spatial columns and temporal snapshots to be subsampled and the creation of a binary mask to implement the subsampling.

The selection process identifies the sets $\mathcal{Y}_{\text{sub}}$ and $\mathcal{T}_{\text{sub}}$, where specific spatial columns are randomly selected to be subsampled in the chosen temporal snapshots. The last snapshot is always included in $\mathcal{T}_{\text{sub}}$, ensuring consistent subsampling at the end of the temporal sequence.

The subsampled dataset $\mathbf{x}_{\text{sub}}$ is obtained according to:
$$
\mathbf{x}_{\text{sub}}(x, y, t) = 
\begin{cases}
0 & \text{if } y \in \mathcal{Y}_{\text{sub}} \text{ and } t \in \mathcal{T}_{\text{sub}} \\
\mathbf{x}(x, y, t) & \text{otherwise}
\end{cases}
$$

Following the subsampling, both the original and subsampled datasets are reshaped to $(n_t, n_x \cdot n_y)$ format for subsequent processing. Normalization is then performed using a single Min-Max Scaler to maintain consistent scaling across all datasets:
$$
\mathbf{X}_{\text{norm}} = \frac{\mathbf{X} - \min(\mathbf{X}_{\text{train}})}{\max(\mathbf{X}_{\text{train}}) - \min(\mathbf{X}_{\text{train}})}
$$
where the scaler is fitted exclusively on the training portion of the data and then applied uniformly to all datasets to prevent data leakage.

The sensor placement process differs between the datasets. For \textbf{SST} and \textbf{qmax} datasets, a preliminary filtering step is applied to remove locations with negligible temporal variation. Let $\mu_p$ be the temporal mean at spatial location $p$ in the subsampled domain of $\mathbf{x}_{\text{sub}} \in \mathbb{R}^{n_t \times n_x \times n_y}$:
$$
\mu_p = \frac{1}{n_t}\sum_{t=1}^{n_t} \mathbf{x}_{\text{sub}}(p, t)
$$

The set of valid locations $\mathcal{P}_{\text{valid}}$ is then defined as:
$$
\mathcal{P}_{\text{valid}} = \{p \in \{1, \ldots, n_x \cdot n_y\} : |\mu_p| > \epsilon\}
$$
where $\epsilon = 10^{-10}$ is the threshold for minimum temporal variation.

The sensor locations are then randomly selected from this filtered set:
$$
\mathcal{S} \subset \mathcal{P}_{\text{valid}}, \quad |\mathcal{S}| = n_{\text{sensors}}
$$

For the Oldroyd-B dataset, the sensor selection is performed directly from the complete subsampled spatial domain:
$$
\mathcal{S} \subset \{1, \ldots, n_x \cdot n_y\}, \quad |\mathcal{S}| = n_{\text{sensors}}
$$

For each sensor location $p \in \mathcal{S}$, sequences of data points are extracted with $l$ lagged time steps:
$$
\mathbf{x}_{\text{sub},p}^{(t)} = [\mathbf{x}_{\text{sub},p,t}, \mathbf{x}_{\text{sub},p,t+1}, \ldots, \mathbf{x}_{\text{sub},p,t+l-1}] \in \mathbb{R}^l
$$

where $t \in \{1, \ldots, n_t-l+1\}$ represents the initial time index of the sequence, and $l$ is the number of time lags. The complete set of sequences for all sensors forms a tensor $\mathbf{X}_{\text{sub}} \in \mathbb{R}^{(n_t-l+1) \times l \times n_{\text{sensors}}}$.

The normalized and sequenced data is then divided into training (70\%), validation (20\%), and testing (10\%) subsets based on the temporal dimension:
\begin{align*}
n_{\text{train}} &= 0.7 \times (n_t - l) \\
n_{\text{val}} &= 0.2 \times (n_t - l) \\
n_{\text{test}} &= 0.1 \times (n_t - l)
\end{align*}

The final datasets are structured as:
\begin{align*}
\mathcal{D}_{\text{train}} &= \{(\mathbf{x}_{\text{sub},p}^{(t)}, \mathbf{x}_{\text{sub},p}^{(t+l)})\}_{t \in \mathcal{I}_{\text{train}}} \\
\mathcal{D}_{\text{val}} &= \{(\mathbf{x}_{\text{sub},p}^{(t)}, \mathbf{x}_{\text{sub},p}^{(t+l)})\}_{t \in \mathcal{I}_{\text{val}}} \\
\mathcal{D}_{\text{test}} &= \{(\mathbf{x}_{\text{sub},p}^{(t)}, \mathbf{x}_{\text{ori},p}^{(t+l)})\}_{t \in \mathcal{I}_{\text{test}}}
\end{align*}

where $\mathbf{x}_{\text{sub},p}^{(t)}$ represents the input lagged sequence from subsampled data at sensor location $p$ starting at time $t$, and $\mathbf{x}_{\text{ori},p}^{(t+l)}$ represents the target from the original (ground truth) data at time step $t+l$. This structured approach ensures that all data is consistently normalized using the same scale. The training and validation use subsampled data for both input and target. Testing properly evaluates reconstruction by comparing subsampled inputs against original data targets and no information leakage occurs between the training, validation, and test sets.

For the \texttt{TURB-Rot} dataset, the data preparation followed the same systematic pipeline described above, with particular attention to the characteristics of turbulent velocity fields. The raw velocity component data (e.g., $v_y$) from multiple HDF5 files were concatenated along the temporal axis to form a comprehensive spatiotemporal array. This array was normalized to zero mean and unit variance to ensure numerical stability during training. The data was then transposed to the $(n_x, n_y, n_t)$ format and truncated to a fixed number of time steps to maintain consistency across experiments. Subsampling was performed by randomly selecting spatial columns and temporal snapshots, always including the final snapshot to preserve temporal coverage. Sensor locations were chosen randomly from the entire spatial domain, without the need for preliminary filtering, reflecting the spatially rich and dynamic nature of turbulent flows. The resulting lagged sequences were constructed as described previously, and the data was split into training, validation, and test sets along the temporal dimension. This approach ensured that the \texttt{TURB-Rot} dataset was processed in a manner fully compatible with the other datasets, enabling fair and consistent evaluation of model performance across diverse spatiotemporal regimes.

Finally, the prepared datasets were encapsulated into \texttt{TimeSeriesDataset} objects, facilitating efficient batching and loading during the training and evaluation phases. This structured approach to data preparation ensured that the models were trained and evaluated on high-quality, well-organized data, thereby enhancing their ability to generalize and accurately reconstruct spatiotemporal dynamics. All results, including model parameters and performance metrics, are systematically saved for reproducibility and further analysis.

\subsection{Viscoelastic Fluid Reconstruction}

The Oldroyd-B model is governed by the following dimensionless equations \cite{OISHI2023}, which integrate the mass and momentum conservation laws with a viscoelastic constitutive relation:

\begin{align*}
\nabla \cdot \mathbf{u} &= 0, \\
\frac{\partial \mathbf{u}}{\partial t} + \mathbf{u} \cdot \nabla \mathbf{u} &= -\nabla p + \frac{\beta}{\text{Re}} \nabla^2 \mathbf{u} + \frac{1}{\text{Re}} \nabla \cdot \mathbf{\tau} + \mathbf{f}, \\
\frac{\partial \mathbf{C}}{\partial t} + (\mathbf{u} \cdot \nabla) \mathbf{C} &= (\nabla \mathbf{u}) \mathbf{C} + \mathbf{C} (\nabla \mathbf{u})^\top \mathbf{T} - \frac{1}{\text{Wi}} (\mathbf{C} - \mathbf{I}), \\
\mathbf{\tau} &= (1 - \beta) \frac{1}{\text{Wi}} (\mathbf{C} - \mathbf{I}),
\end{align*}

where $ \mathbf{u} $ is the fluid velocity, $ p $ is the pressure, $ \mathbf{C} $ is the conformation tensor, $ \mathbf{\tau} $ is the extra stress tensor, and $ \mathbf{f} $ represents external forces. The dimensionless numbers involved are the Reynolds number ($ \text{Re} $), the Weissenberg number ($ \text{Wi} $), and the viscosity ratio ($ \beta $), defined as:

\begin{align*}
\text{Re} &= \frac{\rho UL}{\eta_s + \eta_p}, \\
\text{Wi} &= \frac{\lambda_p U}{L}, \\
\beta &= \frac{\eta_s}{\eta_s + \eta_p},
\end{align*}

where $ \rho $ is the fluid density, $ U $ and $ L $ are characteristic velocity and length scales, $ \eta_s $ and $ \eta_p $ are the solvent and polymer viscosities, respectively, and $ \lambda_p $ is the relaxation time.

Although the Oldroyd-B model can exhibit mathematical singularities under idealized extensive flow conditions, these do not necessarily translate to unrealistic physical behaviors in practical applications. By focusing on the reconstruction of $ \text{tr}(\mathbf{C}) $, we aim to gain a nuanced understanding of the fluid's viscoelastic response under various flow regimes.


Figure~\ref{fig:oldroyd_nosub}a illustrates the final state of a numerical simulation of a viscoelastic fluid governed by the Oldroyd-B framework, which models materials exhibiting both viscous and elastic properties, such as polymers, certain food products, and biological fluids like blood. Unlike simple fluids, viscoelastic fluids demonstrate a hybrid response, displaying characteristics of both liquids and solids. Specifically, Figure~\ref{fig:oldroyd_nosub}a  presents the final time slice of the simulation, visualizing the trace of the conformation tensor, $ \mathrm{tr}(\mathbf{C}) $. This trace serves as a key indicator of elastic stress accumulation across the spatial grid defined by the X and Y axes. The color scale, ranging from black to yellow, highlights regions with high conformation tensor trace values, indicating areas where elastic stress or fluid velocity peaks are most pronounced.

The visual pattern reveals curved structures and intersections, characteristic of viscoelastic materials under shear and deformation, which do not appear in purely viscous fluids. Here, the trace of the conformation tensor indicates the fluid's complex response to external forces, showing accumulation and relaxation of stress over time.

\subsubsection{SHRED Recovering}

The reconstruction of the spatiotemporal dynamics associated with the Oldroyd-B model was performed using the \textbf{SHRED} method. The same hyperparameters listed in Table~\ref{tab:configurations} were employed. The input dataset is a three-dimensional tensor
$$
\mathbf{x}(t) \in \mathbb{R}^{128 \times 128 \times 1433},
$$
which represents the trace of the conformation tensor, $\operatorname{tr}(\mathbf{C})$, with spatial dimensions $128 \times 128$ and 1433 time steps. Measurements are collected by $n_s = 1$ sensor distributed across the spatial domain, yielding the observation $\mathbf{y}(t)$. In our model, a history of $l = 20$ previous time steps,
$$
\{\mathbf{y}(t)\}_{t=t_{\text{current}}-l}^{t_{\text{current}}},
$$
is used to predict the next state of the system. The dataset was partitioned into training ($70\%$ of snapshots), validation ($20\%$), and testing ($10\%$) sets.

The \textbf{SHRED} neural network architecture consists of an LSTM network with 128 hidden units in 1 layer, defining its representation capacity. This is followed by two fully connected layers with $l_1 = 300$ and $l_2 = 400$ neurons, where $l_1$ learns intermediate representations and $l_2$ maps these representations to the final output $\mathbf{x}(t_{\text{current}})$.

\begin{figure}
     \centering
     \begin{subfigure}[b]{0.95\textwidth}
         \centering
         \includegraphics[width=\textwidth]{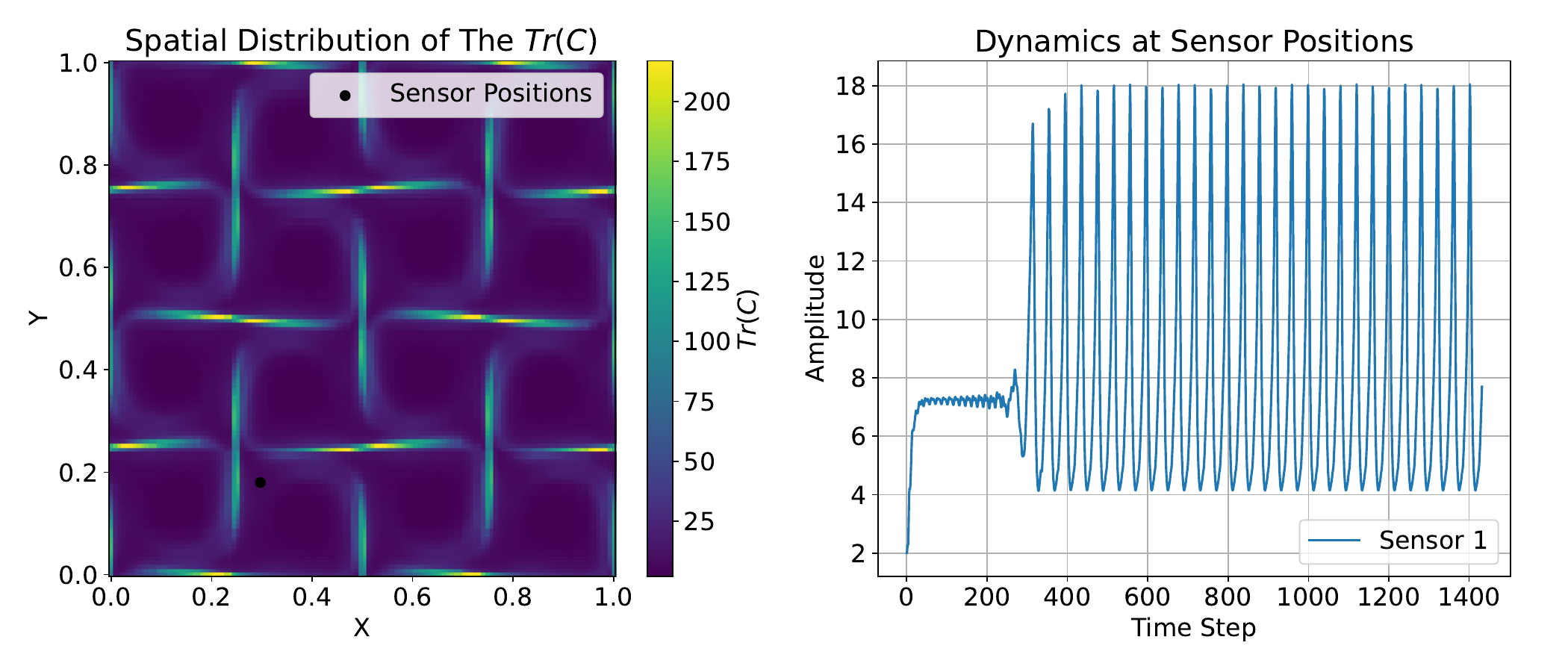}
         \caption{Temporal dynamics.}
         \label{fig:shred_dinan_3}
     \end{subfigure}
     \hfill
     \begin{subfigure}[b]{0.47\textwidth}
         \centering
         \includegraphics[width=\textwidth]{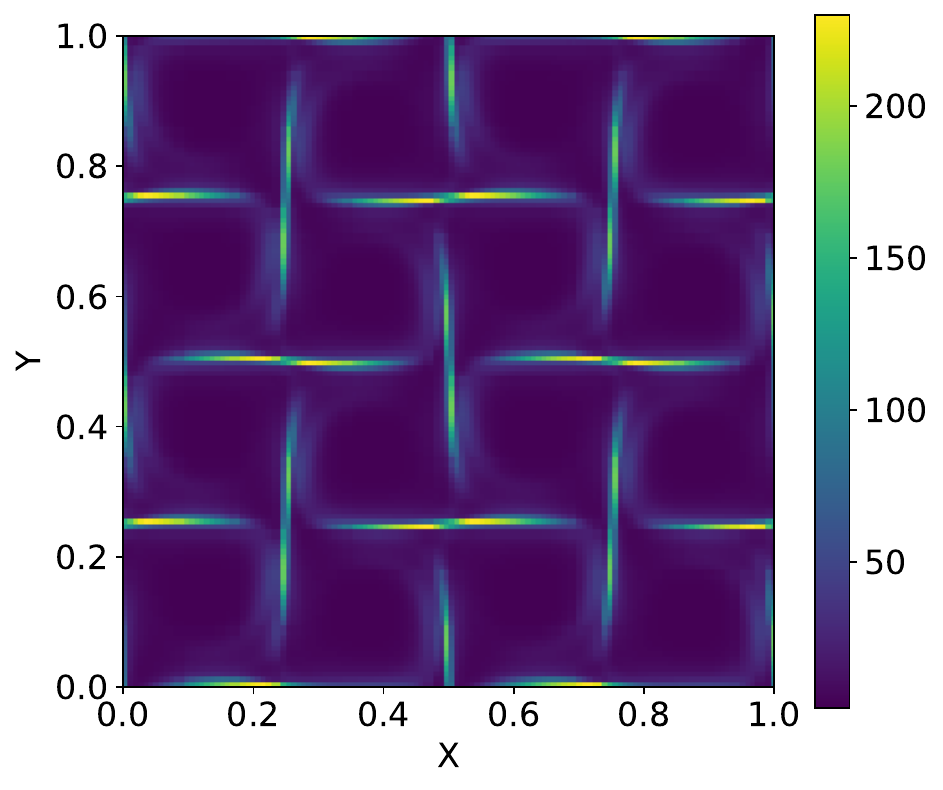}
         \caption{Reconstruction using \textbf{SHRED}.}
         \label{fig:shred_rec_comparison}
     \end{subfigure}
        \caption{Temporal dynamics captured by a sensor fixed at a randomly chosen spatial position (a), and (b) the reconstruction of the spatial distribution of $\operatorname{tr}(\mathbf{C})$ using \textbf{SHRED} with 1 randomly positioned sensor.}
        \label{fig:oldroyd_nosub}
\end{figure}

Training was conducted using the AdamW optimizer with a learning rate $\eta = 0.03420$ over 665 epochs. An \textit{early stopping} mechanism with a patience of 10 epochs was implemented to halt training upon stagnation of the validation metric. The final mean squared error (MSE) on the testing set was reported as
$$
\mathcal{L}_{\text{MSE}} = 0.019886702,
$$
(see Equation~\ref{eq:norm2_error}). Figure~\ref{fig:shred_dinan_3} (right) illustrates the temporal evolution captured by a sensor (black dot) at a randomly selected spatial location -Figure~\ref{fig:shred_dinan_3} (left) -  demonstrating the model’s capacity to reconstruct the complex dynamics of the Oldroyd-B system from sparse sensor placement.

Table~\ref{tab:metrics_shred} presents the quantitative metrics for the recovery of the Oldroyd-B model using \textbf{SHRED}. The reported metrics confirm that the reconstruction is of high quality.

\begin{table}[h]
\centering
\caption{Quantitative metrics obtained in recovering the Oldroyd-B model using \textbf{SHRED} (no subsampled data).}
\label{tab:metrics_shred}
\begin{tabular}{@{}lcc@{}}
\toprule
\textbf{Metric} & \textbf{Value} & \textbf{Ideal} \\
\midrule
\textbf{Metrics (Lower is Better)} & & \\
\quad MSE (Last Snapshot)           & 0.0198867023 & 0 \\
\quad Normalized Error (Last Snapshot) & 0.0205947589 & 0 \\
\quad LPIPS                         & 0.0003287824 & 0 \\
\quad Normalized Error (Mean)       & 0.0192046333 & 0 \\
\midrule
\textbf{Metrics (Higher is Better)} & & \\
\quad SSIM (Last Snapshot)          & 0.9995278347  & 1 \\
\quad PSNR (Last Snapshot) (dB)     & 49.18985781   & Higher is better \\
\quad SSIM (Mean)                   & 0.9994509201  & 1 \\
\quad PSNR (Mean) (dB)              & 49.85812042   & Higher is better \\
\bottomrule
\end{tabular}
\end{table}

Figure~\ref{fig:shred_rec_comparison} shows the reconstruction of the spatial distribution of $\operatorname{tr}(\mathbf{C})$ using \textbf{SHRED} with one randomly positioned sensor. The figure clearly demonstrates that both the spatial patterns and the amplitude variations of the conformation tensor are well preserved.

\subsubsection{Comparison between \textbf{CS-SHRED} and \textbf{SHRED} for Subsampled Data}

A comparative analysis was carried out using temporally \textit{subsampled} multivariate time series acquired by sensors that were \textit{randomly} and \textit{sparsely} positioned throughout the domain, in order to evaluate the ability of \textbf{CS-SHRED} and \textbf{SHRED} to recover the underlying spatiotemporal dynamics.
Table~\ref{tab:configurations} summarizes the configurations for both models, with hyperparameters optimized using the Optuna framework.

\begin{table}[h]
\centering
\caption{Configurations of the \textbf{CS-SHRED} and \textbf{SHRED} models.}
\label{tab:configurations}
\begin{tabular}{@{}lcc@{}}
\toprule
\textbf{Parameter}                        & \textbf{CS-SHRED} & \textbf{SHRED} \\ \midrule
Hidden Size                               & 256               & 128            \\
Hidden Layers                             & 2                 & 1              \\
Batch Size                                & 512               & 128            \\
Learning Rate ($\eta$)                   & 0.00506           & 0.03420        \\
L2 Regularization ($\lambda_{\text{L2}}$) & 0.74010           & -       \\
L1 Regularization ($\lambda_{\text{L1}}$) & 0.00314           & -        \\
SNR Regularization ($\lambda_{\text{SNR}}$) & 0.01183          & -        \\
$l_1$ Parameter                         & 300               & 300            \\
$l_2$ Parameter                         & 400               & 400            \\
Number of Lags                            & 10                & 20             \\
Number of Sensors                         & 1                 & 1              \\
Number of Epochs                          & 1497              & 665            \\
Epoch Step                                & 50                & 50             \\
Seed                                      & 915               & 915            \\
Verbose                                   & True              & True           \\
Patience                                  & 10                & 10             \\
Number of Subsampled Columns              & 115               & 115            \\
Number of Subsampled Snapshots            & 1146              & 1146           \\ \bottomrule
\end{tabular}
\end{table}

For \textbf{CS-SHRED}, the network comprises two hidden layers with 256 neurons each, trained using a batch size of 512 and a learning rate of 0.00506. Regularization is enforced with coefficients $\lambda_{\text{L1}} = 0.00314$, $\lambda_{\text{L2}} = 0.74010$, and an SNR regularization term $\lambda_{\text{SNR}} = 0.01183$. Conversely, the \textbf{SHRED} model features a single hidden layer with 128 neurons, a smaller batch size of 128, and a higher learning rate of 0.03420. Its regularization parameters are $\lambda_{\text{L1}} = 0.00665$, $\lambda_{\text{L2}} = 0.15933$, and $\lambda_{\text{SNR}} = 0.04275$.

\begin{figure}
     \centering
     \begin{subfigure}[b]{0.95\textwidth}
         \centering
         \includegraphics[width=\textwidth]{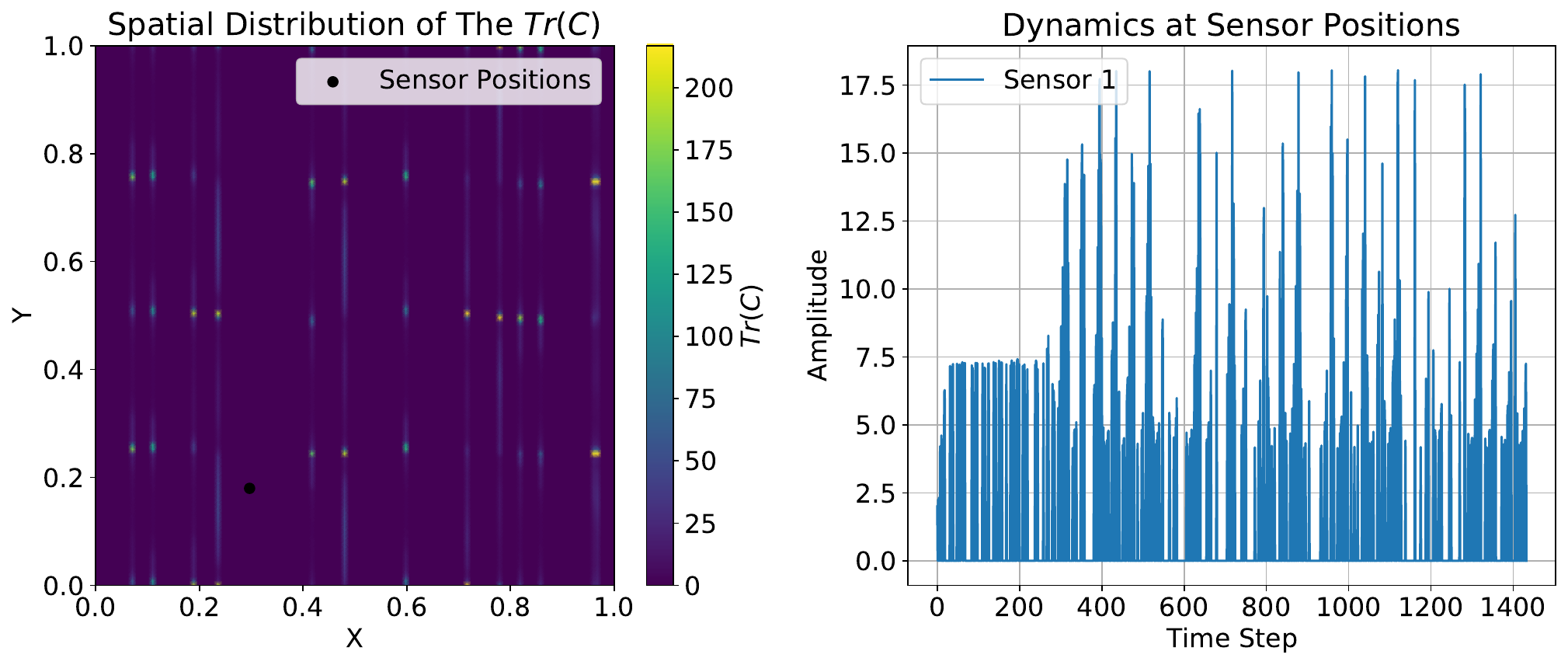}
         \caption{Temporal dynamics.}
         \label{fig:cs_shred_oldroyd}
     \end{subfigure}
     \hfill
     \begin{subfigure}[b]{0.45\textwidth}
         \centering
         \includegraphics[width=\textwidth]{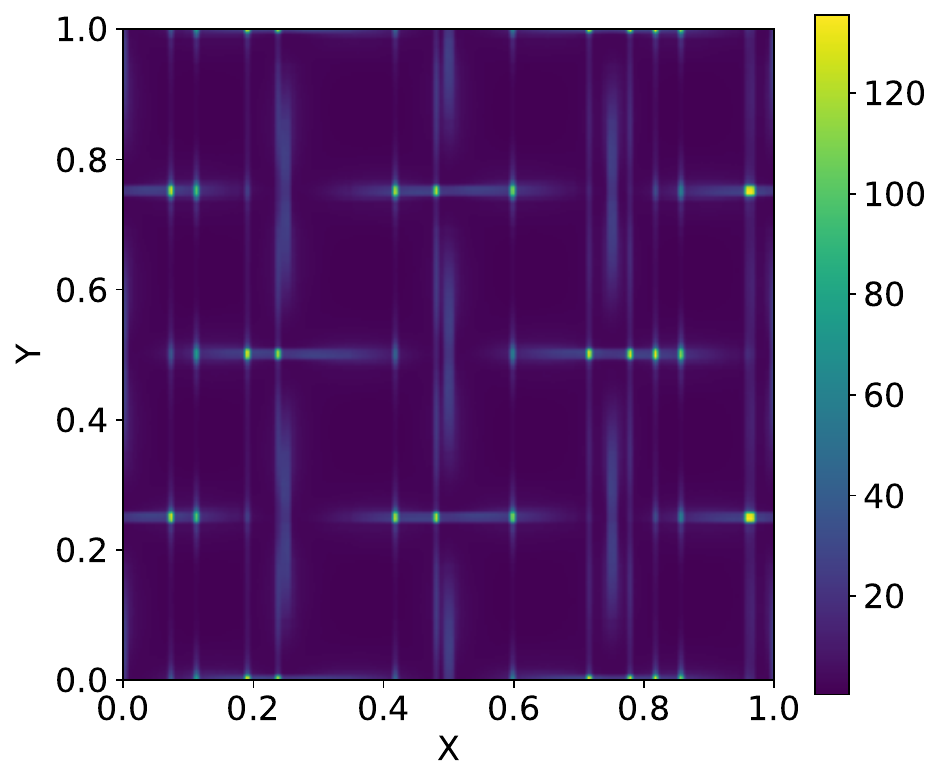}
         \caption{Reconstruction using \textbf{SHRED}.}
         \label{fig:shred_subsampled}
     \end{subfigure}
     \hfill
     \begin{subfigure}[b]{0.45\textwidth}
         \centering
         \includegraphics[width=\textwidth]{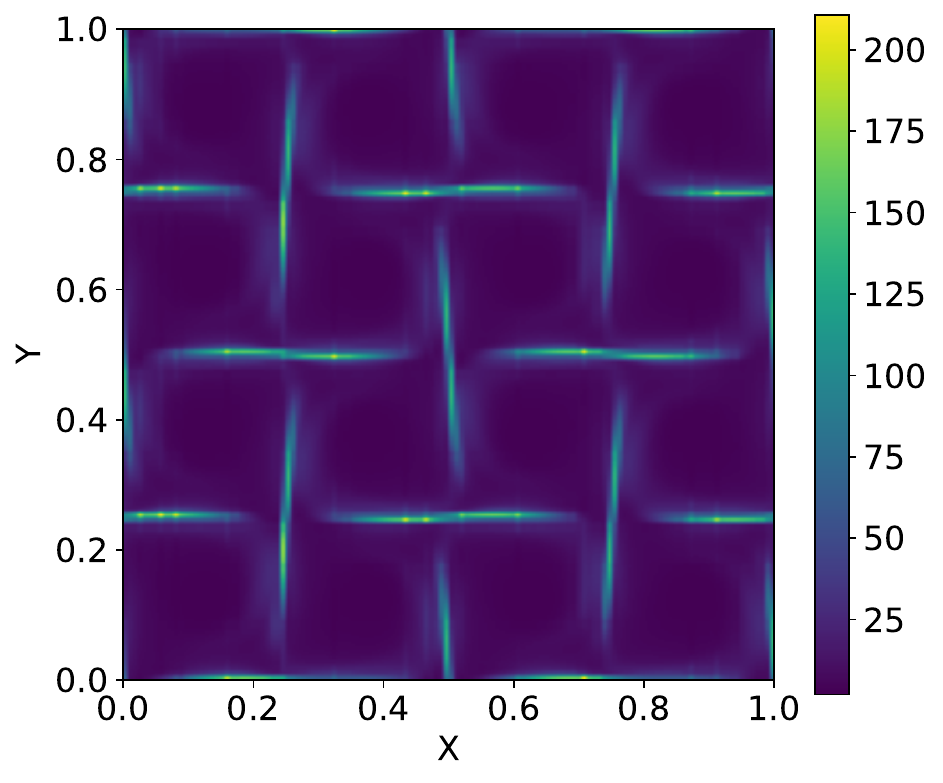}
         \caption{Reconstruction using \textbf{CS-SHRED}.}
         \label{fig:cs-shred_subsampled}
     \end{subfigure}
        \caption{Temporal dynamics captured by a sensor fixed at a randomly chosen spatial position (a), and comparison of the reconstruction of the spatial distribution of $\operatorname{tr}(\mathbf{C})$ using (b) \textbf{SHRED} and (c) \textbf{CS-SHRED} with 1 randomly positioned sensor. Subsampling removed 90\% of the spatial columns (retaining 10\%) and 80\% of the temporal snapshots (preserving 20\%).}
        \label{fig:oldroyd_sub}
\end{figure}

To simulate scenarios with incomplete data, subsampling was applied both spatially and temporally. Spatially, 90\% of the data columns were randomly removed, retaining only 10\% of the columns. Temporally, 80\% of the snapshots were eliminated, preserving only 20\% of the temporal data. Figure~\ref{fig:cs_shred_oldroyd} displays the temporal dynamics captured by a sensor at a randomly selected spatial position under this subsampling scheme.

Figures~\ref{fig:cs-shred_subsampled} and \ref{fig:shred_subsampled} compare the reconstructions of the last snapshot of $\operatorname{tr}(\mathbf{C})$ for \textbf{CS-SHRED} and \textbf{SHRED}, respectively. The corresponding quantitative performance metrics are summarized in Table~\ref{tab:metrics}.

\begin{table}[h]
\centering
\caption{Quantitative metrics comparing \textbf{CS-SHRED} and \textbf{SHRED} methods for subsampled data.}
\label{tab:metrics}
\begin{tabular}{@{}lccc@{}}
\toprule
\textbf{Metric}               & \textbf{CS-SHRED} & \textbf{SHRED} & \textbf{Ideal Value} \\ \midrule
\textbf{Metrics (Lower is Better)} & & & \\
\quad Normalized Error              & 0.29077            & 0.58104        & 0 \\
\quad LPIPS                         & 0.04708            & 0.13045        & 0 \\
\midrule
\textbf{Metrics (Higher is Better)} & & & \\
\quad SSIM (Last Snapshot)          & 0.95199            & 0.72995        & 1 \\
\quad PSNR (Last Snapshot) (dB)     & 27.47              & 20.12          & Higher is better \\
\quad SSIM (Mean)                   & 0.92472            & 0.73260        & 1 \\
\quad PSNR (Mean) (dB)              & 27.12              & 22.27          & Higher is better \\ \bottomrule
\end{tabular}
\end{table}

The reconstruction performance indicates a clear advantage for \textbf{CS-SHRED} over \textbf{SHRED}. For the last snapshot, \textbf{CS-SHRED} achieved an SSIM of 0.952 compared to 0.730 for \textbf{SHRED}, suggesting that \textbf{CS-SHRED} maintains superior structural fidelity and minimizes distortions. In addition, the PSNR for \textbf{CS-SHRED} was 27.47 dB versus 20.12 dB for \textbf{SHRED}, which implies that \textbf{CS-SHRED} introduces less noise and better captures the intensity variations of the conformation tensor. This is further corroborated by the lower MSE and LPIPS values for \textbf{CS-SHRED}, confirming its enhanced precision and perceptual quality.

Figure~\ref{fig:cs-shred_subsampled} compares the \textbf{CS-SHRED} reconstruction of the last snapshot with the available subsampled data. Despite the significant data loss—only 10\% of the spatial columns and 20\% of the temporal snapshots are retained—\textbf{CS-SHRED} is able to effectively reconstruct the missing information. This demonstrates the robustness of the method in handling incomplete datasets.

Maintaining high fidelity in reconstructing the conformation tensor is critical for accurately representing the viscoelastic properties and dynamic behavior of fluids governed by the Oldroyd-B equations. The accurate reconstruction facilitates detailed analysis of internal stresses and deformations, leading to improved predictions in industrial and scientific simulations.

Overall, Figure~\ref{fig:oldroyd_sub} provides a comprehensive visual comparison of the reconstructions obtained using both \textbf{CS-SHRED} and \textbf{SHRED} methods alongside the original spatial distribution of $\operatorname{tr}(\mathbf{C})$.

\subsubsection{Computational Performance Analysis}

Following the performance comparison, we conducted a detailed computational performance analysis to understand the trade-offs between superior reconstruction quality and computational efficiency. All experiments were conducted on a system equipped with an Intel Core i7 processor, 16 GB of RAM, and an NVIDIA GeForce GTX 1650 GPU with 4 GB of VRAM, running Ubuntu 22.04 LTS with CUDA 12.9 support. It is important to remember that all hyperparameters were obtained with the aid of the Optuna framework, which explored comparable hyperparameter spaces for both models. For the specific dataset $\text{Tr}(\mathbf{C})$ of the analyzed Oldroyd-B model, \textbf{CS-SHRED} converged to a more complex configuration due to its superior validation performance, though this pattern may not generalize to other datasets with different characteristics.

As shown in Table~\ref{tab:execution_times}, the \textbf{CS-SHRED} model requires 3.0 times more execution time than \textbf{SHRED}, with training being the primary computational bottleneck in both models. Specifically, \textbf{CS-SHRED} took 159.82 seconds (2.66 minutes) compared to SHRED's 53.07 seconds (0.88 minutes) for complete execution. This increased computational cost is directly related to the more complex architecture of \textbf{CS-SHRED}, which features additional hidden layers and larger batch sizes. However, the superior reconstruction quality achieved by \textbf{CS-SHRED}, as evidenced by the metrics in Table~\ref{tab:metrics} where it achieved substantially better SSIM (0.952 vs 0.730), PSNR (27.47 dB vs 20.12 dB), and lower normalized error (0.291 vs 0.581), is primarily due to the methodology employed rather than the computational complexity, as demonstrated by the fact that in other datasets CS-SHRED achieved superior results even with simpler architectures (see Section \ref{sec:sst_model_config}).


\begin{table}[h]
\centering
\caption{Execution times by operation (in seconds)}
\label{tab:execution_times}
\begin{tabular}{|l|c|c|c|}
\hline
\textbf{Operation} & \textbf{SHRED} & \textbf{CS-SHRED} & \textbf{Difference (\%)} \\
\hline
train\_and\_validate\_model & 51.62 & 158.32 & +206.7 \\
prepare\_datasets & 1.09 & 1.04 & -4.6 \\
subsample & 0.20 & 0.28 & +40.0 \\
evaluate\_model & 0.16 & 0.18 & +12.5 \\
\hline
\textbf{Total Time} & \textbf{53.07} & \textbf{159.82} & \textbf{+201.2} \\
\hline
\end{tabular}
\end{table}

Table~\ref{tab:memory_usage} presents the detailed memory consumption for each operation of each model. Both models exhibit similar peak memory usage, with \textbf{CS-SHRED} using 828.01 MB compared to \textbf{SHRED}'s 785.37 MB, representing only a 5.4\% increase despite the significantly more complex architecture. The memory consumption during training shows that \textbf{CS-SHRED} uses 495.25 MB compared to \textbf{SHRED}'s 457.62 MB, representing only an 8.2\% increase. This suggests that the enhanced reconstruction performance of \textbf{CS-SHRED} comes primarily at the cost of computational time rather than memory requirements, making it an attractive option for applications where reconstruction accuracy is paramount and sufficient computational time is available.

\begin{table}[h]
\centering
\caption{Memory usage by operation (in MB)}
\label{tab:memory_usage}
\begin{tabular}{|l|c|c|c|}
\hline
\textbf{Operation} & \textbf{SHRED} & \textbf{CS-SHRED} & \textbf{Difference (\%)} \\
\hline
train\_and\_validate\_model & 457.62 & 495.25 & +8.2 \\
prepare\_datasets & 122.45 & 131.33 & +7.2 \\
subsample & 179.45 & 179.55 & +0.06 \\
evaluate\_model & 25.85 & 21.88 & -15.4 \\
\hline
\textbf{Peak Total} & \textbf{785.37} & \textbf{828.01} & \textbf{+5.4} \\
\hline
\end{tabular}
\end{table}

Table~\ref{tab:model_complexity} presents the detailed architectural complexity characteristics. The \textbf{CS-SHRED} architecture features 81\% more parameters (332,582 vs 183,504) and a computational complexity 4.8 times greater in the forward pass (104,120,320 vs 21,626,880 operations). The architectural differences include a larger hidden size (256 vs 128 neurons), additional hidden layers (2 vs 1), and a significantly larger batch size (512 vs 128). These differences, however, are specific to this dataset and, as previously explained, do not explain the superior reconstruction quality, which stems from the methodology itself rather than architectural complexity.

\begin{table}[h]
\centering
\caption{Model complexity characteristics}
\label{tab:model_complexity}
\begin{tabular}{|l|c|c|c|}
\hline
\textbf{Parameter} & \textbf{SHRED} & \textbf{CS-SHRED} & \textbf{Difference (\%)} \\
\hline
Total parameters & 183,504 & 332,582 & +81.2 \\
Forward pass complexity & 21,626,880 & 104,120,320 & +381.4 \\
Memory per batch & 0.42 MB & 2.41 MB & +473.8 \\
Hidden size & 128 & 256 & +100.0 \\
Hidden layers & 1 & 2 & +100.0 \\
Batch size & 128 & 512 & +300.0 \\
Lags & 20 & 10 & -50.0 \\
\hline
\end{tabular}
\end{table}

Table~\ref{tab:computational_efficiency} presents the derived computational efficiency metrics. While \textbf{SHRED} demonstrates higher efficiency in terms of time per parameter (0.28 vs 0.48 ms) and memory usage, \textbf{CS-SHRED} achieves higher operation throughput (657,456 vs 418,847 operations per second) due to its larger batch size. The operations per second metric is calculated at the batch level, where the 4x larger batch size significantly increases global throughput, even though the normalized latency per parameter increases. This trade-off highlights the fundamental difference between the models: \textbf{SHRED} prioritizes computational efficiency, while \textbf{CS-SHRED} prioritizes reconstruction quality, achieving significantly better performance metrics at the cost of computational resources. The GPU utilization during training showed that both models effectively utilized the NVIDIA GeForce GTX 1650, with \textbf{CS-SHRED} maintaining higher sustained GPU utilization due to its larger batch processing requirements.

\begin{table}[h]
\centering
\caption{Computational efficiency metrics}
\label{tab:computational_efficiency}
\begin{tabular}{|l|c|c|c|}
\hline
\textbf{Metric} & \textbf{SHRED} & \textbf{CS-SHRED} & \textbf{Difference (\%)} \\
\hline
Time per parameter (ms) & 0.28 & 0.48 & +71.4 \\
Memory per parameter (kB) & 2.33 & 1.49 & -36.1 \\
Operations per second & 418,847 & 657,456 & +57.0 \\
Memory efficiency (MB/s) & 31.2 & 10.9 & -65.1 \\
\hline
\end{tabular}
\end{table}

The memory efficiency metric (MB/s) reflects the bytes transferred per training time and is not critical since the computational bottleneck limits performance. Considering the time complexity $O(T \times H \times W)$ for data operations and $O(P \times B)$ for training, where $P$ is the number of parameters and $B$ is the batch size, \textbf{CS-SHRED} presents a 7.2x higher training complexity ($O(332,582 \times 512) = O(170.3M)$ vs $O(183,504 \times 128) = O(23.5M)$). This directly explains the significantly higher runtime observed experimentally. However, the efficient parallelization of large batches by the GPU amortizes this theoretical overhead, which explains why the actual overhead (2.7x) is smaller than the FLOPs ratio. This computational investment does not explain the substantial improvements in reconstruction quality, since for other datasets studied (see Section~\ref{sec:sst_model_config}) the hyperparameters chosen by Optuna made \textbf{CS-SHRED} less complex than \textbf{SHRED} while still providing high-fidelity reconstructions.

The computational analysis reveals a trade-off between reconstruction quality and computational efficiency that  must be considered when selecting between models. \textbf{SHRED} offers superior computational efficiency (3.0x faster execution) but with lower reconstruction quality (SSIM: 0.730, PSNR: 20.12 dB). Conversely, \textbf{CS-SHRED} provides superior reconstruction quality (SSIM: 0.952, PSNR: 27.47 dB) at the cost of higher computational requirements (for this specific case), making it ideal for applications where reconstruction accuracy is critical and sufficient computational resources are available. Importantly, the reconstruction results obtained through  \textbf{CS-SHRED} have been systematically superior across all analyzed cases, demonstrating the robustness and effectiveness of the methodology. Both models have similar peak memory usage, with  \textbf{CS-SHRED} requiring only 5.4\% more memory than \textbf{SHRED}, which simplifies deployment considerations.

\subsection{Maximum Specific Humidity (\textbf{qmax})}

In this section, we apply the reconstruction framework introduced in the previous section to the specific humidity \textbf{qmax} dataset. As before, subsampling techniques are employed to mimic realistic data acquisition challenges, such as sensor limitations, environmental variability, and maintenance issues \cite{hart2006environmental,trenberth2013challenges,anderson2008quality}. In particular, we remove 90\% of the data in 30\% of the temporal snapshots (retaining 1,209 out of 1,727 snapshots) and a similar proportion of spatial columns. This subsampling approach is consistent with the strategies discussed in \cite{donoho2006compressed,candes2008introduction,fowler2012compressive}.

\begin{figure}
     \centering
     \begin{subfigure}[b]{0.47\textwidth}
         \centering
         \includegraphics[width=\textwidth]{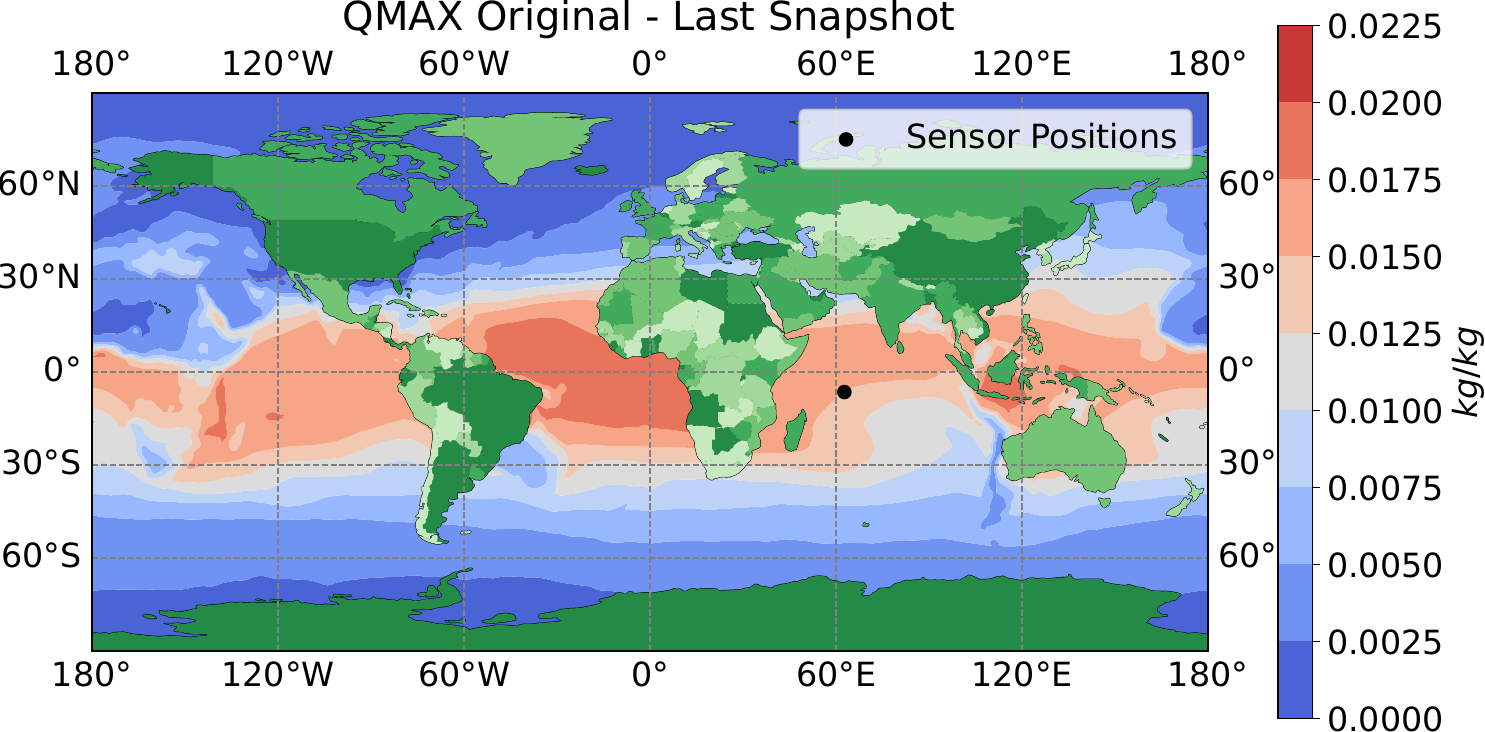}
         \caption{Original distribution}
         \label{fig:qmax_ori}
     \end{subfigure}
     \hfill
     \begin{subfigure}[b]{0.47\textwidth}
         \centering
         \includegraphics[width=\textwidth]{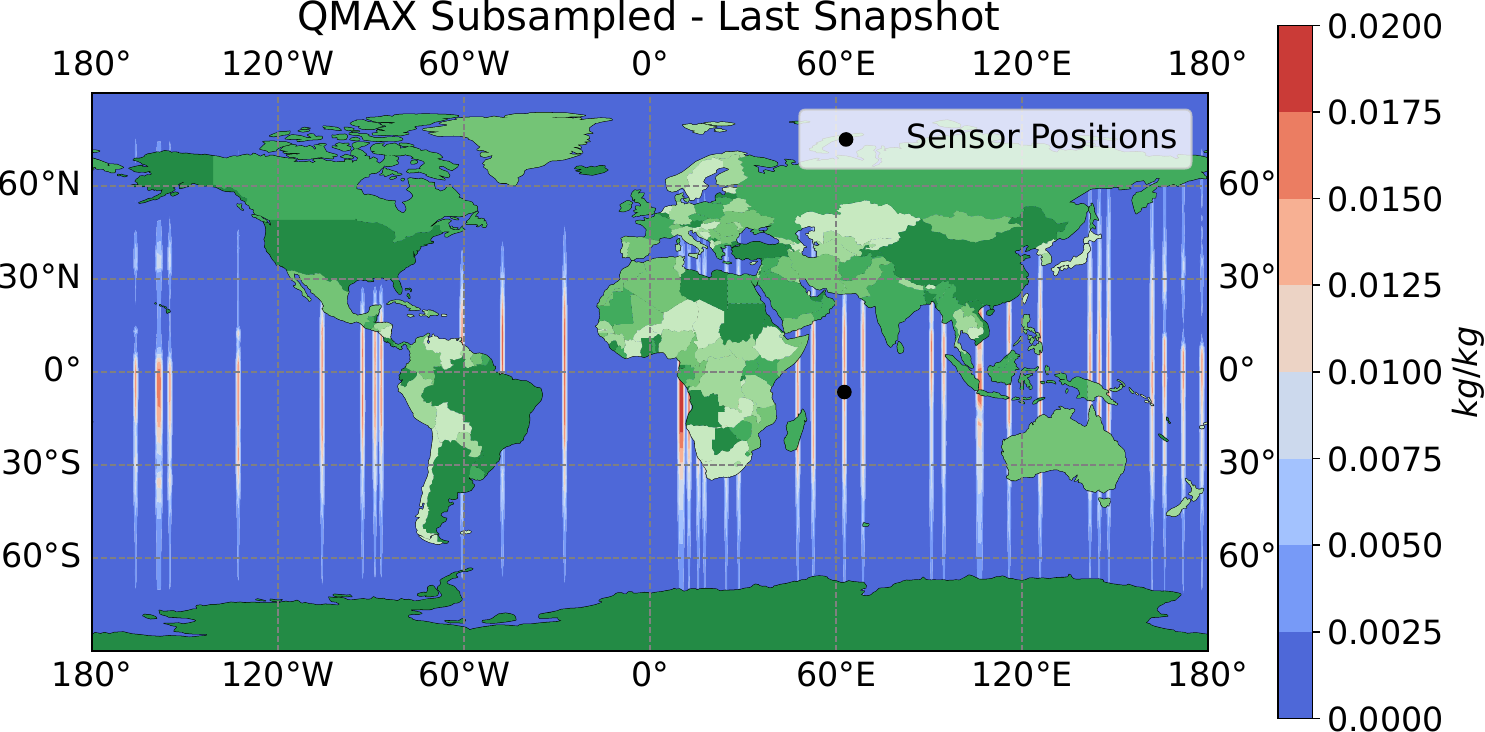}
         \caption{Subsampled distribution}
         \label{fig:qmax_sub}
     \end{subfigure}
     \hfill
     \begin{subfigure}[b]{1\textwidth}
         \centering
         \includegraphics[width=\textwidth]{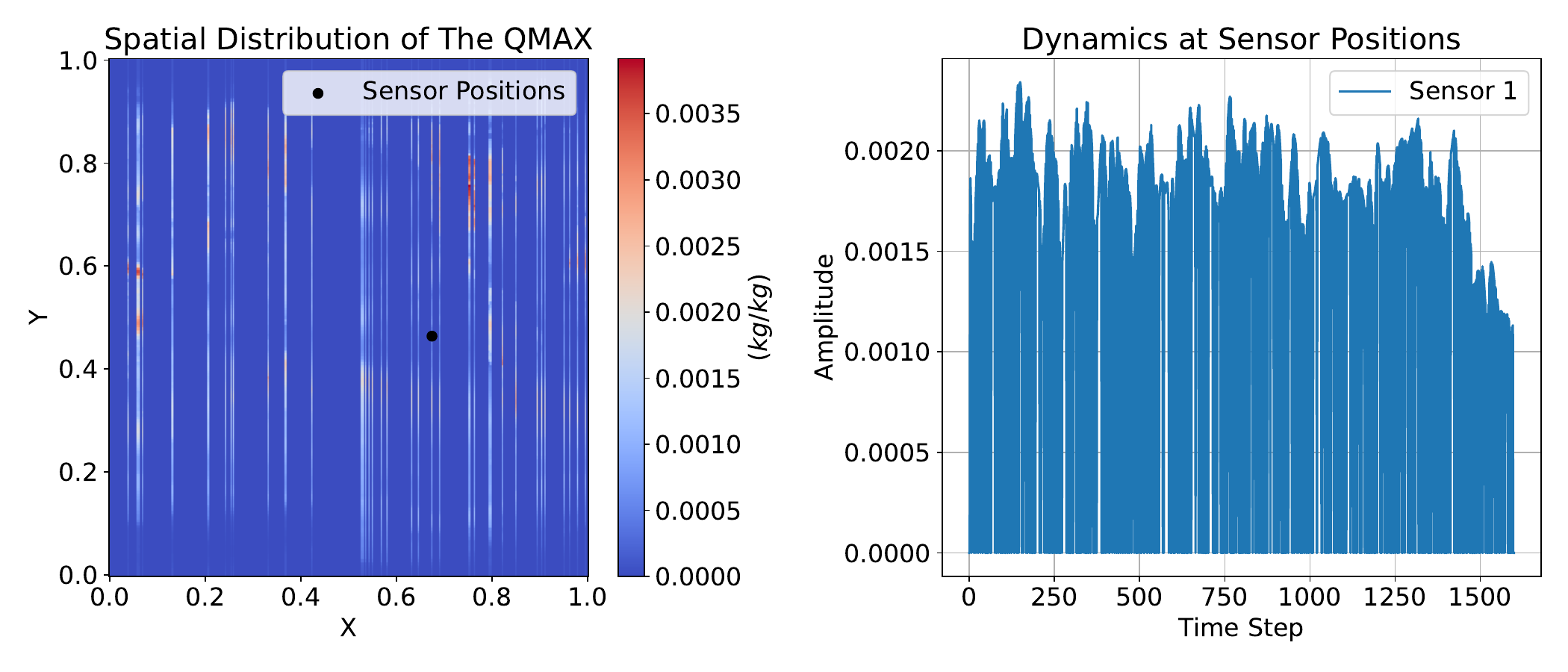}
         \caption{Temporal dynamics}
         \label{fig:qmax_din}
     \end{subfigure}
     \hfill
     \begin{subfigure}[b]{0.47\textwidth}
         \centering
         \includegraphics[width=\textwidth]{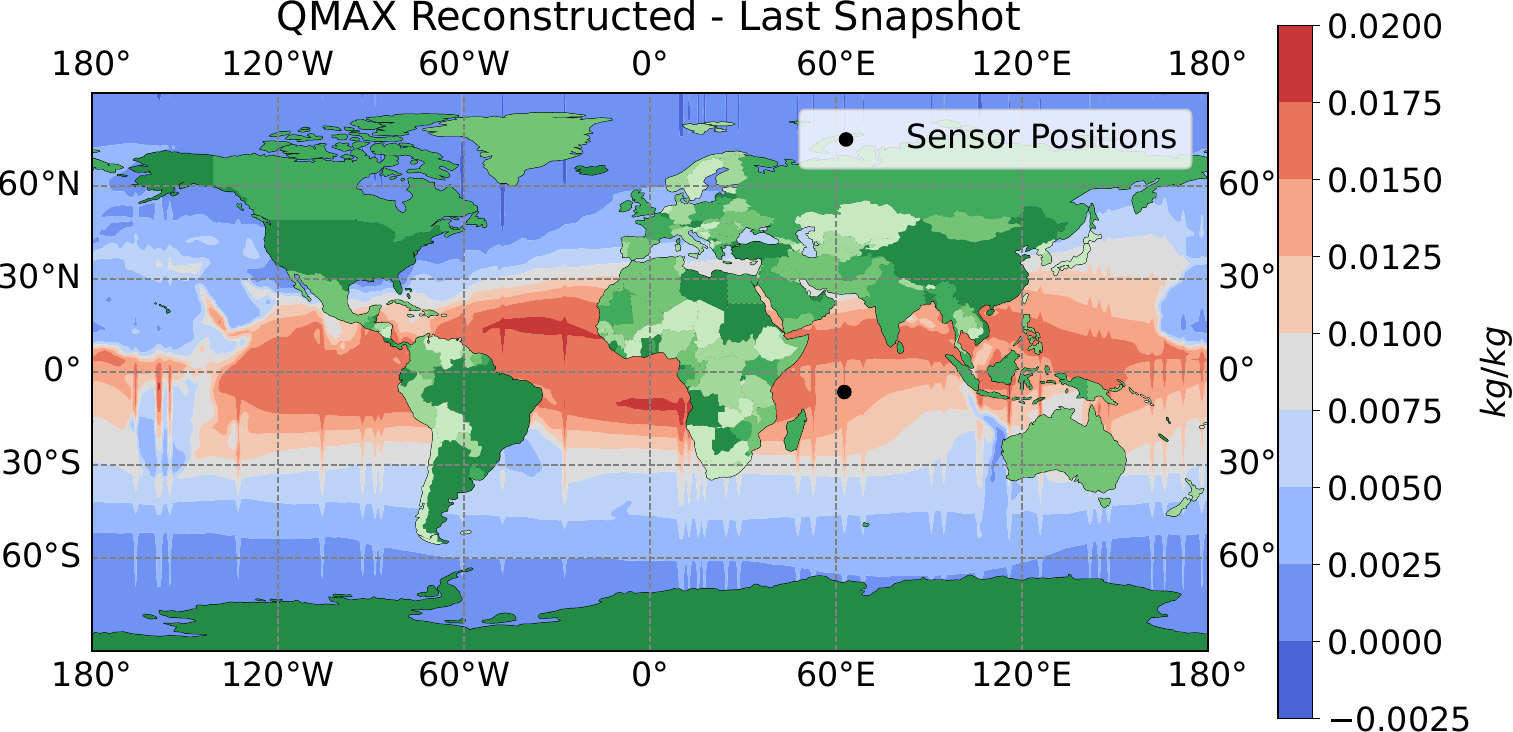}
         \caption{\textbf{SHRED}}
         \label{fig:qmax_sh}
     \end{subfigure}
      \hfill
     \begin{subfigure}[b]{0.47\textwidth}
         \centering
         \includegraphics[width=\textwidth]{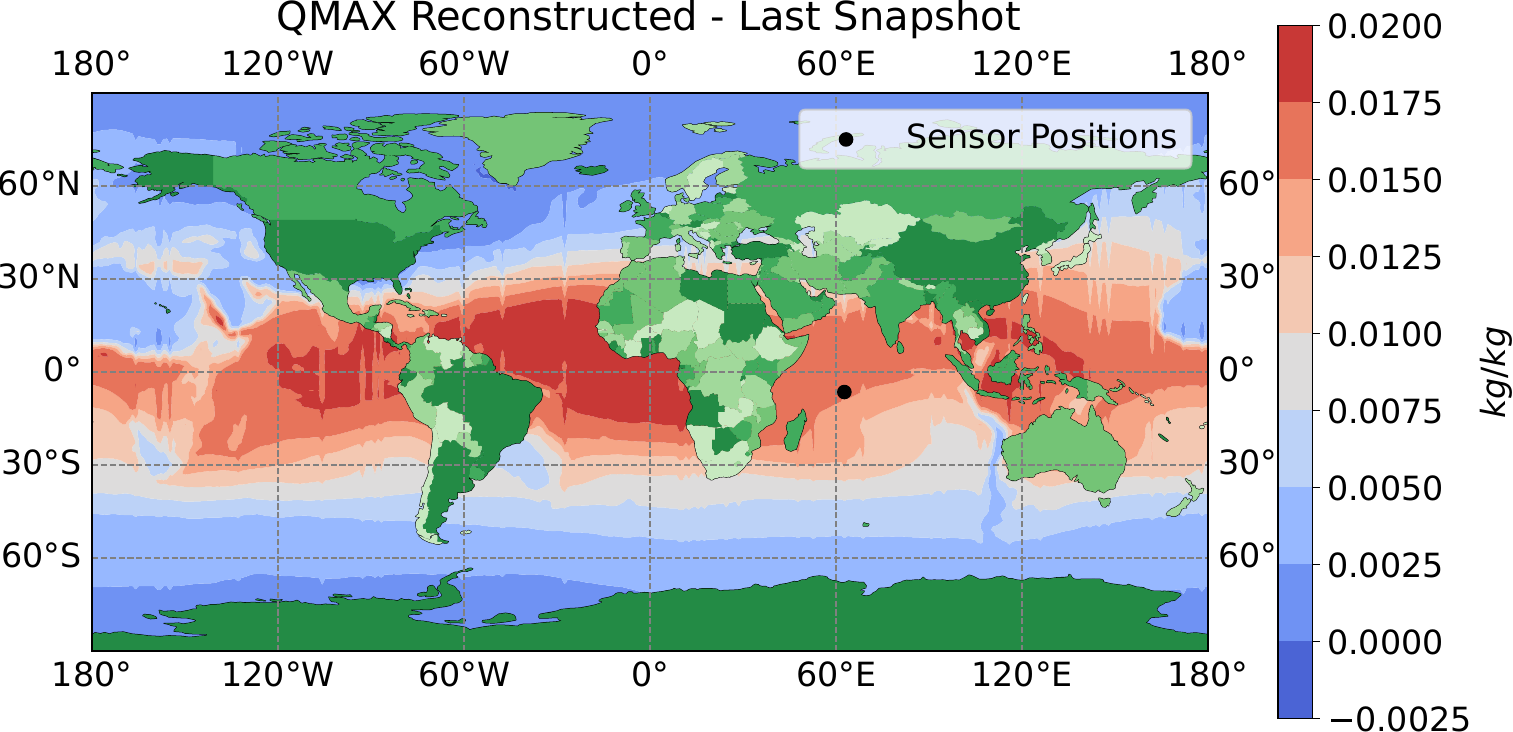}
         \caption{\textbf{CS-SHRED}}
         \label{fig:qmax_cs}
     \end{subfigure}
        \caption{Comparison between \textbf{SHRED} and \textbf{CS-SHRED} for the specific humidity (\textbf{qmax}) data: (a) shows the original data of the last snapshot, (b) shows the respective subsampled data, (c) shows the temporal dynamics captured by a sensor fixed at a randomly chosen spatial position. The resultant reconstructions are shown at (d) for \textbf{SHRED} and (e) for \textbf{CS-SHRED}.}
        \label{fig:qmax_ori_sub_1s}
\end{figure}

Figure~\ref{fig:qmax_ori_sub_1s} shows the original spatial distribution of specific humidity and its subsampled version. In the complete distribution (Figure~\ref{fig:qmax_ori}), the color scale represents humidity concentration (in kg of water per kg of dry air), where red regions indicate higher concentrations and blue regions lower concentrations. Figure~\ref{fig:qmax_sub} presents the corresponding subsampled distribution.

Subsequently, Figure~\ref{fig:qmax_din} illustrates the temporal dynamics recorded by a randomly selected sensor (Sensor 1), emphasizing the challenges posed by amplitude variations in the humidity signal under limited data acquisition conditions.

The reconstruction of the \textbf{qmax} field from subsampled data was performed using both the \textbf{CS-SHRED} and \textbf{SHRED} methods. Their performance was evaluated using quantitative metrics (SSIM, PSNR, normalized error, and LPIPS), as summarized in Table~\ref{tab:metricas_qmax}. Figures~\ref{fig:qmax_cs} and \ref{fig:qmax_sh} present the reconstructed spatial distributions of \textbf{qmax} for each method.

\subsection{Model Configurations for Maximum Specific Humidity Data (\textbf{qmax})}

Table~\ref{tab:configuracoes_qmax} summarizes the configurations for the \textbf{CS-SHRED} and \textbf{SHRED} models optimized via the Optuna framework. As in the previous section, both models are designed to reconstruct the spatiotemporal dynamics from subsampled data, with adjustments made to parameters such as hidden size, learning rate, number of lags, and regularization coefficients to best suit the characteristics of the \textbf{qmax} dataset.

\begin{table}[h]
\centering
\caption{Configurations of the \textbf{CS-SHRED} and \textbf{SHRED} Models for \textbf{qmax} Data}
\label{tab:configuracoes_qmax}
\begin{tabular}{@{}lcc@{}}
\toprule
\textbf{Parameter}                           & \textbf{CS-SHRED}              & \textbf{SHRED}                  \\ \midrule
Hidden Size                                  & 256                            & 512                             \\
Hidden Layers                                & 1                              & 2                               \\
Batch Size                                   & 512                            & 256                             \\
Learning Rate ($\eta$)                      & 0.000377                       & 0.009849                        \\
L2 Regularization ($\lambda_{\text{L2}}$)    & 0.376978                       & -                               \\
L1 Regularization ($\lambda_{\text{L1}}$)    & 0.013111                       & -                               \\
SNR Regularization ($\lambda_{\text{SNR}}$)   & 0.003763                       & -                               \\
$l_1$ Parameter                            & 500                            & 700                             \\
$l_2$ Parameter                            & 600                            & 600                             \\
Number of Lags                               & 24                             & 54                              \\
Number of Sensors                            & 1                              & 1                               \\
Number of Epochs                             & 421                            & 273                             \\
Epoch Step                                   & 50                             & 50                              \\
Seed                                         & 915                            & 915                             \\
Verbose                                      & True                           & True                            \\
Patience                                     & 15                             & 15                              \\
Number of Subsampled Columns                 & 324                            & 324                             \\
Number of Subsampled Snapshots               & 518                            & 518                             \\ \bottomrule
\end{tabular}
\end{table}

For \textbf{CS-SHRED}, a neural network with a single hidden layer (256 neurons) is used with a batch size of 512 and a learning rate of 0.000377. In contrast, \textbf{SHRED} utilizes a deeper architecture with two hidden layers (512 neurons each) and a higher learning rate of 0.009849. The primary difference between the models lies in the number of lags (24 for \textbf{CS-SHRED} versus 54 for \textbf{SHRED}) and the corresponding training epochs.

\subsubsection{Comparison between \textbf{CS-SHRED} and \textbf{SHRED} for Subsampled \textbf{qmax} Data}

Figures~\ref{fig:qmax_cs} and \ref{fig:qmax_sh} display the reconstructed spatial distributions of \textbf{qmax} for the last snapshot using \textbf{CS-SHRED} and \textbf{SHRED}, respectively. Quantitative metrics listed in Table~\ref{tab:metricas_qmax} further quantify the performance of both methods.

\begin{table}[h]
\centering
\caption{Quantitative Metrics Comparing \textbf{CS-SHRED} and \textbf{SHRED} Methods for \textbf{qmax}, with Ideal Ranges}
\label{tab:metricas_qmax}
\begin{tabular}{@{}lccc@{}}
\toprule
\textbf{Metric}                & \textbf{CS-SHRED}    & \textbf{SHRED}      & \textbf{Ideal Value} \\ \midrule
\textbf{Metrics (Lower is Better)} & & & \\
\quad Normalized Error       & 0.09326              & 0.12594              & 0                    \\
\quad LPIPS                  & $7.42 \times 10^{-5}$ & $1.36 \times 10^{-4}$ & 0                    \\ \midrule
\textbf{Metrics (Higher is Better)} & & & \\
\quad Mean SSIM              & 0.91158              & 0.87070              & 1                    \\
\quad Mean PSNR (dB)         & 27.54                & 25.28                & Higher is better     \\
\quad SSIM (Last Snapshot)   & 0.89098              & 0.78202              & 1                    \\
\quad PSNR (Last Snapshot) (dB) & 25.16             & 21.22                & Higher is better     \\ \bottomrule
\end{tabular}
\end{table}

The results indicate that \textbf{CS-SHRED} outperforms \textbf{SHRED} in reconstructing \textbf{qmax}. In particular, \textbf{CS-SHRED} achieves a higher SSIM (0.891 vs. 0.782) and PSNR (25.16 dB vs. 21.22 dB) in the last snapshot, indicating better preservation of structural details and lower reconstruction noise. These improvements are also reflected in the lower normalized error and LPIPS values for \textbf{CS-SHRED}, confirming its superior performance in capturing the essential characteristics of the humidity field despite significant data subsampling.


\subsection{Sea Surface Temperature Analysis (\textbf{SST})}\label{sec:sst_result}

In this section, we apply our reconstruction framework to Sea Surface Temperature (\textbf{SST}) data. As in previous sections, the focus is on recovering the spatiotemporal dynamics from subsampled data. Here, we first describe the model configurations and then compare the performance of the \textbf{CS-SHRED} and \textbf{SHRED} models.

\subsubsection{Model Configurations}\label{sec:sst_model_config}

Table~\ref{tab:configuracoes_temperatura} summarizes the configurations of the \textbf{CS-SHRED} and \textbf{SHRED} models, which were optimized via the Optuna hyperparameter framework. These settings were adjusted specifically for the \textbf{SST} dataset to ensure optimal recovery of its spatiotemporal dynamics.

\begin{table}[h]
\centering
\caption{Configurations of the \textbf{CS-SHRED} and \textbf{SHRED} Models for Sea Surface Temperature  Data}
\label{tab:configuracoes_temperatura}
\begin{tabular}{@{}lcc@{}}
\toprule
\textbf{Parameter}                           & \textbf{CS-SHRED}              & \textbf{SHRED}                  \\ \midrule
Hidden Size                                  & 512                            & 512                             \\
Hidden Layers                                & 2                              & 2                               \\
Batch Size                                   & 64                             & 512                             \\
Learning Rate ($\eta$)                      & $3.08 \times 10^{-4}$         & $3.40 \times 10^{-4}$          \\
L2 Regularization ($\lambda_{\text{L2}}$)    & 0.3220                         & -                               \\
L1 Regularization ($\lambda_{\text{L1}}$)    & 0.0041                         & -                               \\
SNR Regularization ($\lambda_{\text{SNR}}$)   & $1.30 \times 10^{-3}$         & -                               \\
$l_1$ Parameter                            & 500                            & 400                             \\
$l_2$ Parameter                            & 300                            & 300                             \\
Number of Lags                               & 36                             & 24                              \\
Number of Sensors                            & 1                              & 1                               \\
Number of Epochs                             & 397                            & 640                             \\
Epoch Step                                   & 20                             & 20                              \\
Seed                                         & 915                            & 915                             \\
Verbose                                      & True                           & True                            \\
Patience                                     & 15                             & 15                              \\
Number of Subsampled Columns                 & 324                            & 324                             \\
Number of Subsampled Snapshots               & 518                            & 518                             \\ \bottomrule
\end{tabular}
\end{table}

For the \textbf{CS-SHRED} model, a neural network with two hidden layers (each with 512 neurons) is used with a batch size of 64 and a learning rate of $3.08 \times 10^{-4}$. Regularizations are applied with $\lambda_{\text{L1}} = 0.0041$, $\lambda_{\text{L2}} = 0.3220$, and $\lambda_{\text{SNR}} = 1.30 \times 10^{-3}$. In contrast, the \textbf{SHRED} model also employs two hidden layers of 512 neurons but uses a larger batch size (512) and a slightly higher learning rate ($3.40 \times 10^{-4}$). Its regularization coefficients are set to $\lambda_{\text{L1}} = 0.0215$ and $\lambda_{\text{L2}} = 0.0669$ with an SNR term of 0.6396. Both models use $l_1 = 500$ and $l_2 = 300$ for the shallow decoder network, with the key difference being the number of lags (36 for \textbf{CS-SHRED} and 24 for \textbf{SHRED}) and the number of training epochs.

All other parameters (seed, verbosity, patience, and subsampling settings) are kept constant to ensure a fair comparison. In this case, the subsampling procedure retains 10\% of the spatial columns and 30\% of the temporal snapshots (518 snapshots), simulating realistic data acquisition constraints.

\subsubsection{Comparison between \textbf{CS-SHRED} and \textbf{SHRED} for Subsampled \textbf{SST} Data}

\begin{figure}
     \centering
     \begin{subfigure}[b]{0.47\textwidth}
         \centering
         \includegraphics[width=\textwidth]{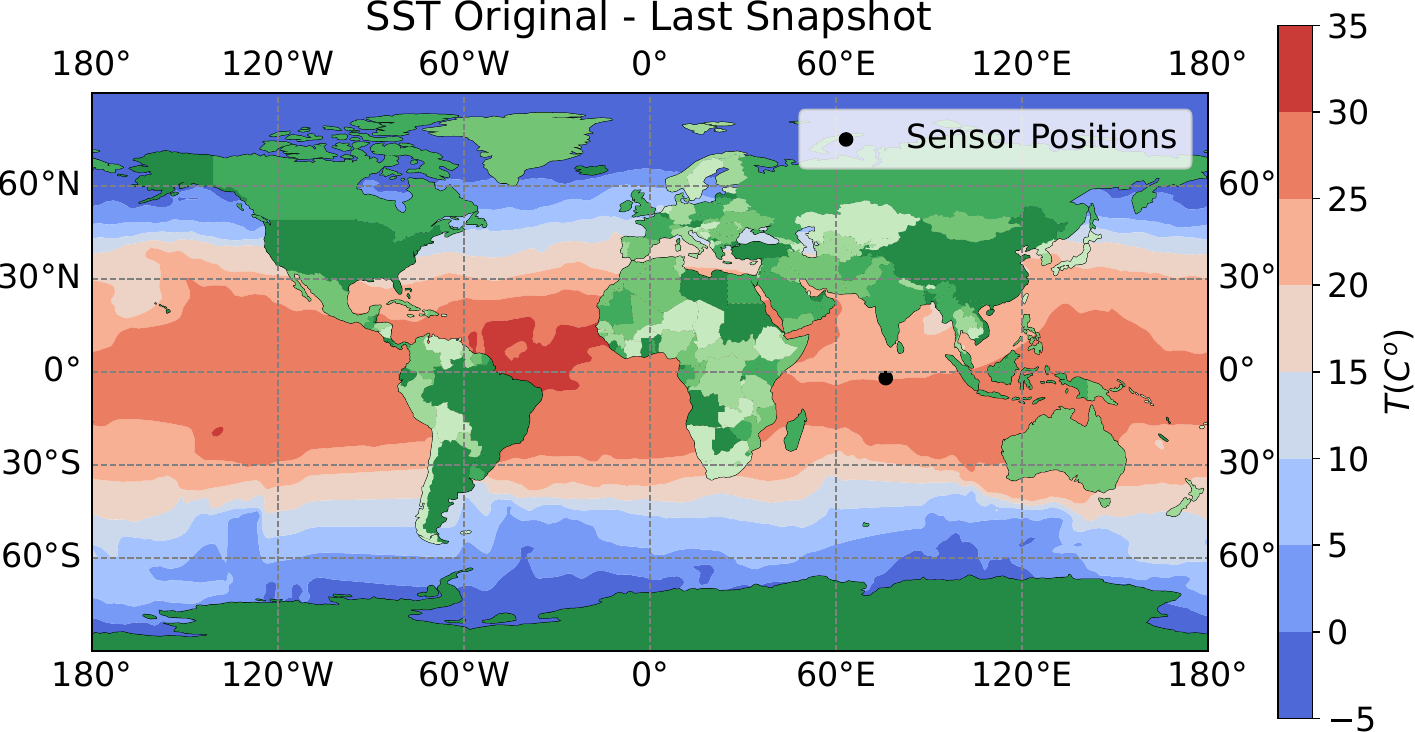}
         \caption{Original distribution}
         \label{fig:sst_ori}
     \end{subfigure}
     \hfill
     \begin{subfigure}[b]{0.47\textwidth}
         \centering
         \includegraphics[width=\textwidth]{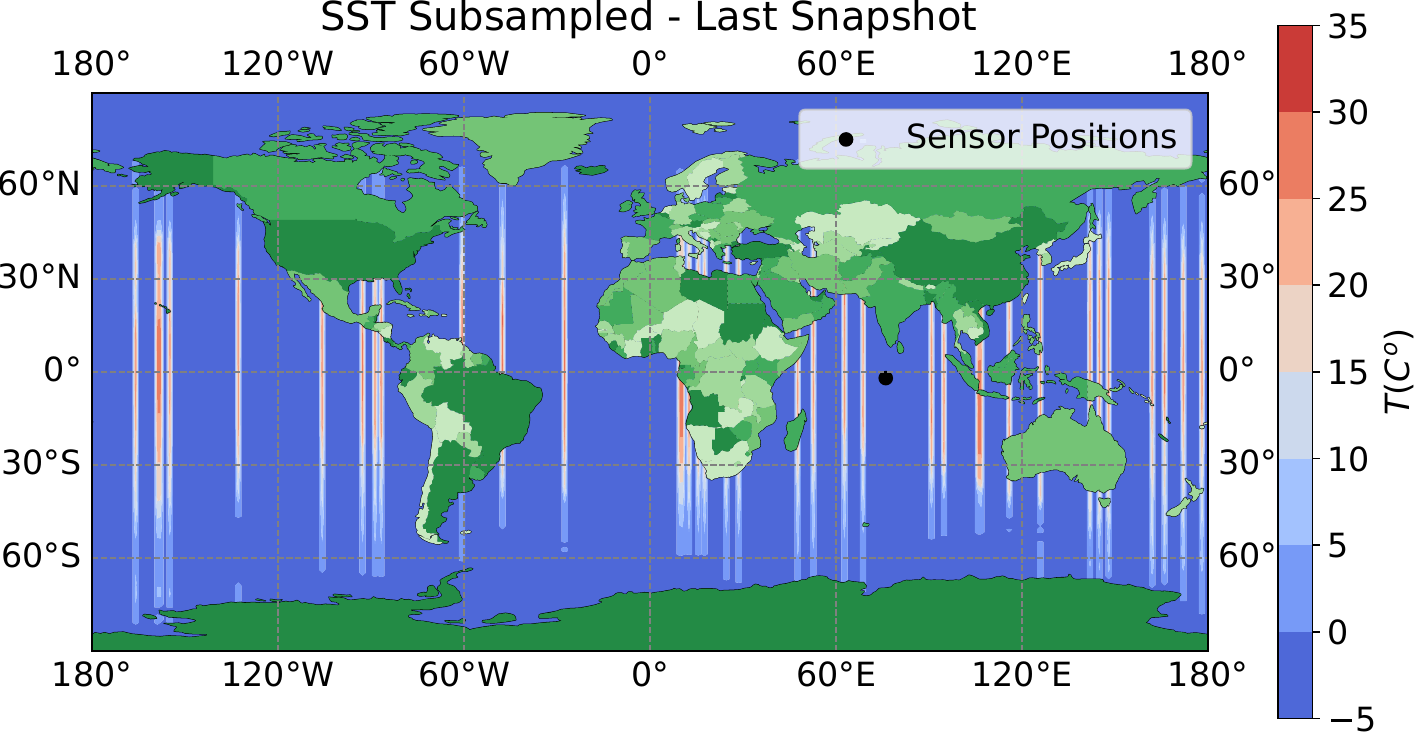}
         \caption{Subsampled distribution}
         \label{fig:sst_sub}
     \end{subfigure}
     \hfill
     \begin{subfigure}[b]{1\textwidth}
         \centering
         \includegraphics[width=\textwidth]{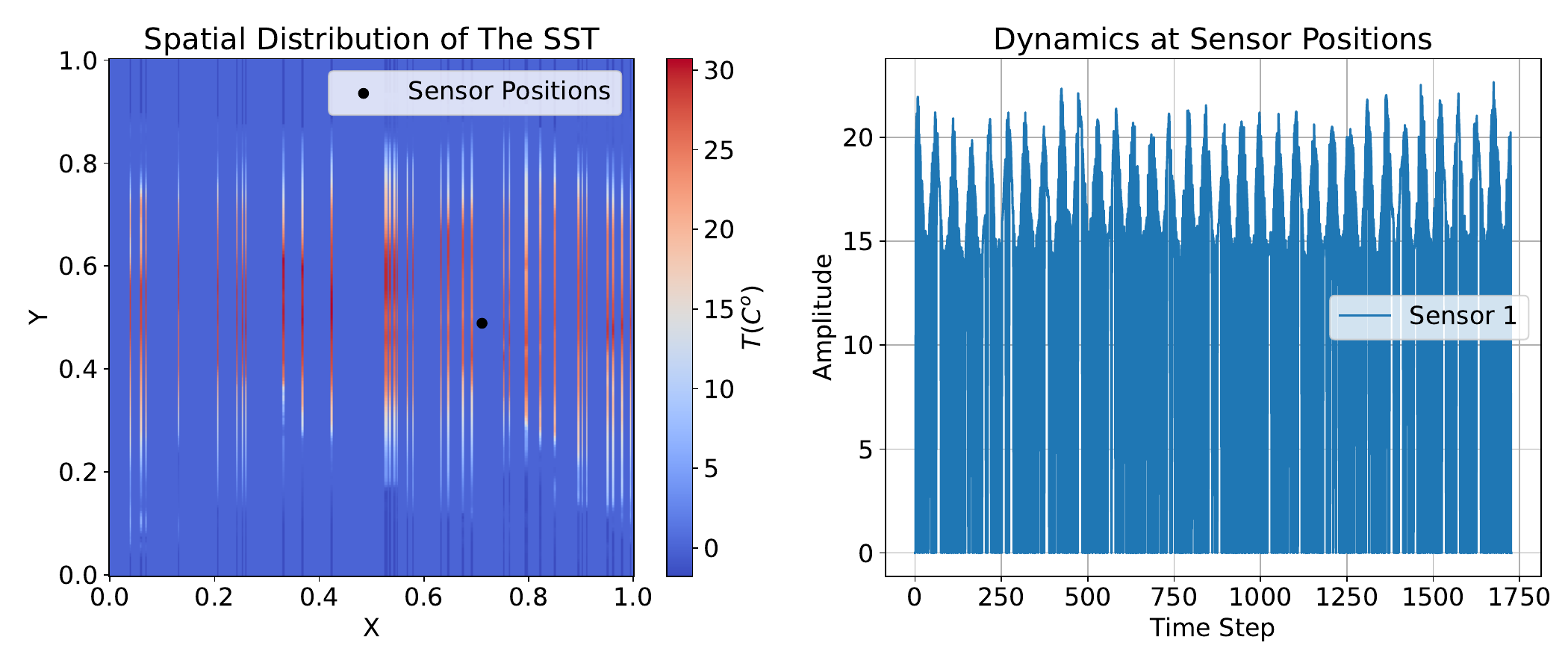}
         \caption{Temporal dynamics}
         \label{fig:sst_din}
     \end{subfigure}
     \hfill
     \begin{subfigure}[b]{0.47\textwidth}
         \centering
         \includegraphics[width=\textwidth]{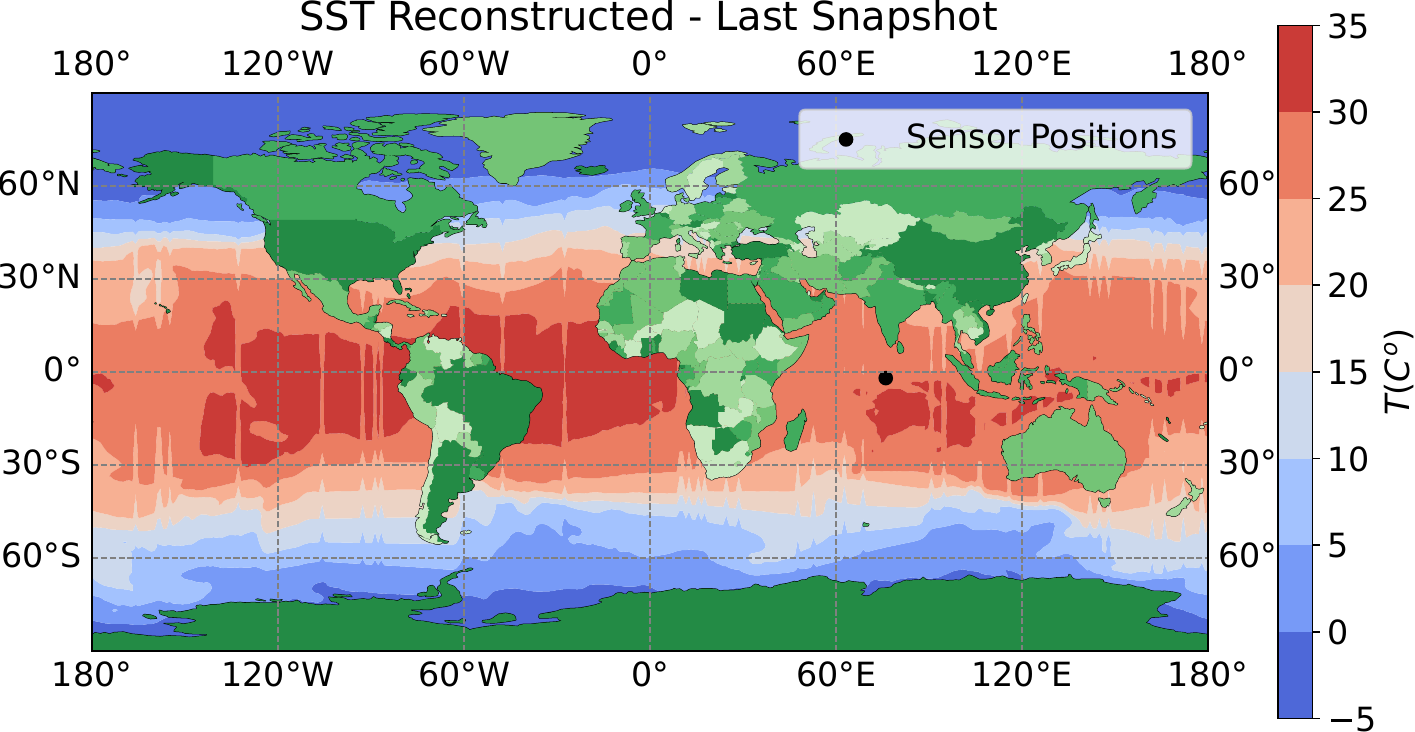}
         \caption{\textbf{SHRED}}
         \label{fig:sst_sh}
     \end{subfigure}
      \hfill
     \begin{subfigure}[b]{0.47\textwidth}
         \centering
         \includegraphics[width=\textwidth]{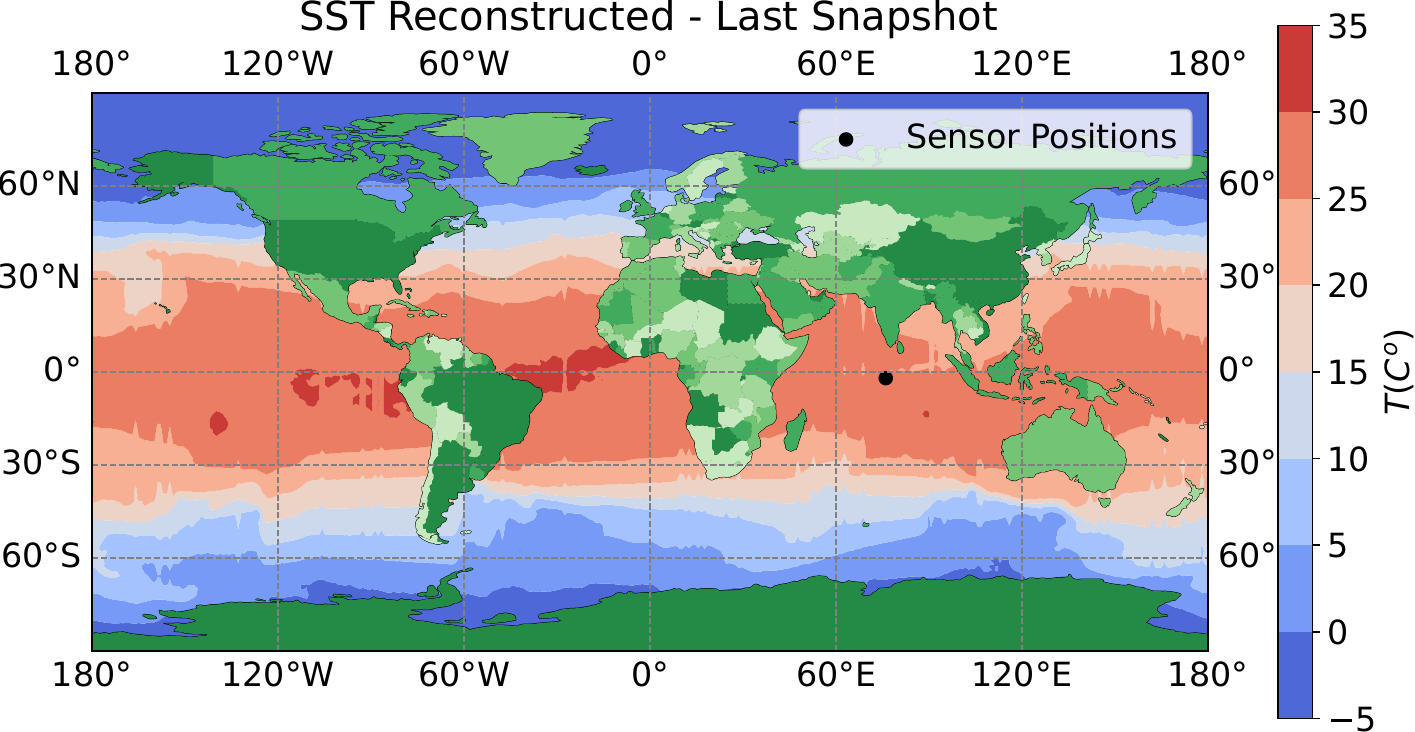}
         \caption{\textbf{CS-SHRED}}
         \label{fig:sst_cs}
     \end{subfigure}
        \caption{Comparison between \textbf{SHRED} and \textbf{CS-SHRED} for the Sea Surface Temperature  data: (a) shows the original data of the last snapshot, (b) shows the respective subsampled data, (c) shows the temporal dynamics captured by a sensor fixed at a randomly chosen spatial position. The resultant reconstructions are shown at (d) for \textbf{SHRED} and (e) for \textbf{CS-SHRED}.}
        \label{fig:sst_ori_sub}
\end{figure}

Figures~\ref{fig:sst_ori} and \ref{fig:sst_sub} display the complete and subsampled spatial distributions of \textbf{SST}, respectively. In Figure~\ref{fig:sst_ori}, the color scale represents temperature in degrees Celsius (with red indicating higher and blue indicating lower temperatures), while Figure~\ref{fig:sst_sub} shows the corresponding subsampled version where 90\% of the data is removed in 30\% of the temporal snapshots.

Figure~\ref{fig:sst_din} illustrates the temporal dynamics captured by a randomly selected sensor (Sensor 1), highlighting the amplitude variations in the temperature signal over time, which underscores the challenges of reconstructing \textbf{SST} dynamics under limited data conditions.

The reconstruction results are presented in Figures~\ref{fig:sst_cs} and \ref{fig:sst_sh} for \textbf{CS-SHRED} and \textbf{SHRED}, respectively. Table~\ref{tab:metricas_sst} summarizes the corresponding quantitative metrics.

\begin{table}[h]
\centering
\caption{Quantitative Metrics Comparing \textbf{CS-SHRED} and \textbf{SHRED} Methods for \textbf{SST}, with Ideal Ranges}
\label{tab:metricas_sst}
\begin{tabular}{@{}lccc@{}}
\toprule
\textbf{Metric}                & \textbf{CS-SHRED}    & \textbf{SHRED}      & \textbf{Ideal Value} \\ \midrule
\textbf{Metrics (Lower is Better)} & & & \\
\quad Normalized Error       & 0.1589               & 0.1709              & 0                    \\
\quad LPIPS                  & 0.2472               & 0.2586              & 0                    \\ \midrule
\textbf{Metrics (Higher is Better)} & & & \\
\quad Mean SSIM              & 0.7468               & 0.7267              & 1                    \\
\quad Mean PSNR (dB)         & 21.95                & 21.19               & Higher is better     \\
\quad SSIM (Last Snapshot)   & 0.8717               & 0.7905              & 1                    \\
\quad PSNR (Last Snapshot) (dB) & 28.81             & 22.58               & Higher is better     \\ \bottomrule
\end{tabular}
\end{table}

The quantitative metrics reveal that \textbf{CS-SHRED} provides a more accurate and faithful reconstruction of the \textbf{SST} field. In particular, \textbf{CS-SHRED} attains a higher SSIM (0.8717 vs. 0.7905) and PSNR (28.81~dB vs. 22.58~dB) in the last snapshot, indicating superior preservation of structural details and lower noise. These improvements are further supported by lower normalized error and LPIPS values.

\subsection{Rotating Turbulent Flow Reconstruction}
\label{sec:rotating_flow}

In this section, we investigate a rotating incompressible flow using data from the \texttt{TURB-Rot} database \cite{Biferale2020TURBRotAL}. The simulation was conducted with a fully dealiased, parallel, three-dimensional pseudospectral code on a $256^3$ grid in a triply periodic domain of size $L = 2\pi$. A second-order Adams--Bashforth method was employed for time-stepping, while the viscous (or hyperviscous) term was integrated implicitly. In the rotating reference frame, the incompressible Navier--Stokes equations are given by:
\begin{align}
\frac{\partial \mathbf{u}}{\partial t} + (\mathbf{u} \cdot \nabla)\mathbf{u} + 2\,\boldsymbol{\Omega} \times \mathbf{u} 
&= -\nabla p + \nu\,\nabla^2 \mathbf{u} + \alpha\,\nabla^{-2}\mathbf{u} + \mathbf{f}, \\
\nabla \cdot \mathbf{u} &= 0,
\label{eq:rot_ns}
\end{align}
where $\mathbf{u}(\mathbf{x},t)$ is the velocity field, $p(\mathbf{x},t)$ is the modified pressure, and $\boldsymbol{\Omega} = \Omega\,\hat{\mathbf{x}}_3$ is the angular velocity vector about the $\hat{\mathbf{x}}_3$ axis. The Coriolis force appears as $2\,\boldsymbol{\Omega} \times \mathbf{u}$. Here, $\nu$ is the kinematic viscosity and the hypo-viscous term $\alpha\,\nabla^{-2}\mathbf{u}$ (with $\alpha = 0.1$) is added to prevent large-scale condensates \cite{Alexakis_2018}. Energy is injected by a Gaussian, delta-correlated forcing $\mathbf{f}$ in a narrow wavenumber band around $\lvert \mathbf{k} \rvert \approx 4$. With $\Omega = 8$, the Rossby number is defined as
$$
\text{Ro} = \frac{\sqrt{E}}{k_f\,\Omega} \approx 0.1,
$$
with $E$ representing the flow kinetic energy and $k_f = 4$ the forcing wavenumber. To further enhance small-scale damping, the standard Laplacian $\nabla^2$ in the viscous term is replaced by a hyperviscous operator $\nabla^4$ with $\nu = 1.6 \times 10^{-6}$.

\subsubsection{Data Extraction and Assembly for Flow Analysis}
\label{sec:data_extraction}

For our analysis, we selected a subset of the \texttt{TURB-Rot} dataset corresponding to a statistically steady regime. Although data for all three velocity components $(u_x,\,u_y,\,u_z)$ were available, we focus primarily on the $u_y$ component for initial analysis. The extracted data were organized into three-dimensional arrays (in $\mathrm{x, y, z}$) from which we computed basic statistics (minimum, maximum, mean) and the velocity magnitude to assess global energy levels.

\begin{figure}[htbp]
    \centering
    \includegraphics[width=0.98\textwidth]{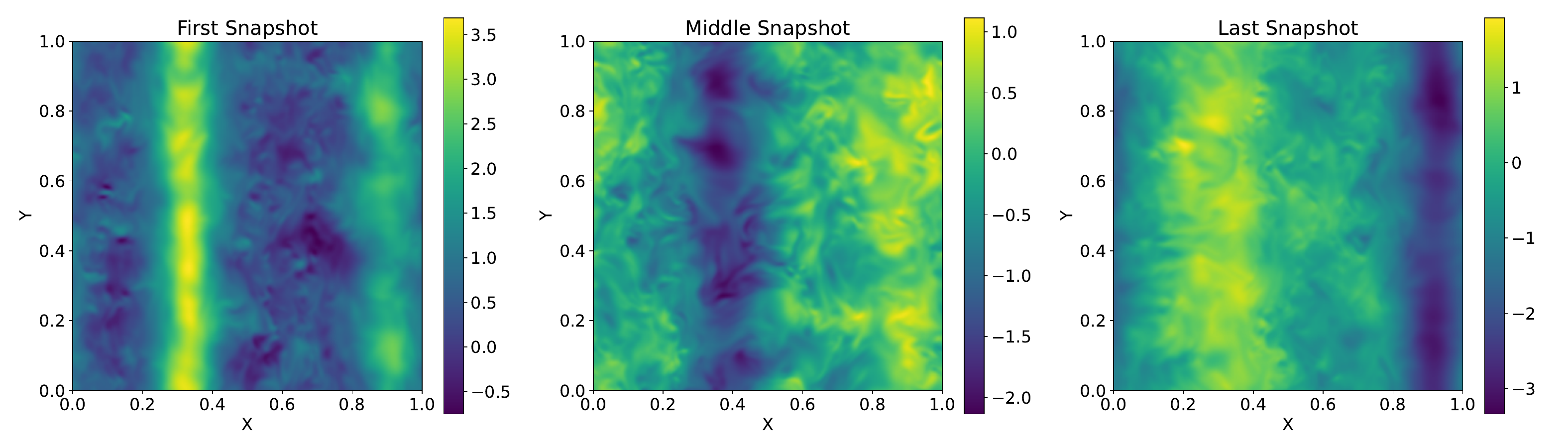}
    \caption{Visualization of the rotating turbulent flow focusing on the $u_y$ velocity component. The extracted $z$-plane reveals coherent vortex-like structures.}
    \label{fig:turb_vy_combined}
\end{figure}

We have also computed empirical probability-density functions (PDFs) for each component by normalised histograms (not shown for brevity). Their heavier tails for \(u_y\) corroborate the larger intermittency of the transverse dynamics and further justify the choice of \(u_y\) as the primary target in the reconstruction experiments. Figure~\ref{fig:turb_vy_combined} shows an example visualization of the $u_y$ component, highlighting these vortex-like features. Finally, the subset was saved as a \texttt{NumPy} archive to ensure reproducibility and ease of post-processing.

\subsubsection{Model Configurations for Rotating Turbulent Flow}\label{sec:rotating_flow_results}

Table~\ref{tab:config_rotating_flow} lists the hyperparameter configurations for the \textbf{CS-SHRED} and \textbf{SHRED} models applied to the rotating turbulent flow dataset. Both models are implemented with three hidden layers; however, \textbf{CS-SHRED} uses a larger hidden size (256 vs. 128 in \textbf{SHRED}). Both models employ 15 lags and 5 sensors, ensuring direct comparability. Notable differences include the batch size (32 for \textbf{CS-SHRED} vs. 128 for \textbf{SHRED}), and the inclusion of L1, L2, and SNR regularizations in \textbf{CS-SHRED}. In addition, \textbf{CS-SHRED} converged in 913 epochs while \textbf{SHRED} required 1871 epochs; both models use a step schedule to reduce the learning rate at regular intervals.

\begin{table}[h]
\centering
\caption{Model configurations for \textbf{CS-SHRED} and \textbf{SHRED} applied to the rotating turbulent flow (\texttt{TURB-Rot}).}
\label{tab:config_rotating_flow}
\begin{tabular}{@{}lcc@{}}
\toprule
\textbf{Parameter}         & \textbf{CS-SHRED}                 & \textbf{SHRED}                   \\ \midrule
Hidden Size                & 256                               & 128                              \\
Hidden Layers              & 3                                 & 3                                \\
Batch Size                 & 32                                & 128                              \\
Learning Rate ($\eta$)     & $1.49 \times 10^{-3}$           & $5.88 \times 10^{-3}$            \\
L2 Regularization          & 0.2513                            & -                                \\
L1 Regularization          & 0.0091                            & -                                \\
SNR Regularization         & 0.8552                            & -                                \\
Dropout                    & 0.0111                            & 0.0104                           \\
$l_1$ (Parameter)     & 400                               & 300                              \\
$l_2$ (Parameter)     & 400                               & 500                              \\
Number of Lags             & 15                                & 15                               \\
Number of Sensors          & 5                                 & 5                                \\
Number of Epochs           & 913                               & 1871                             \\
Epoch Step                 & 28                                & 23                               \\
Seed                       & 915                               & 915                              \\
Verbose                    & True                              & True                             \\
Patience                   & 15                                & 15                               \\
Subsampled Columns         & 77                                & 77                               \\
Subsampled Snapshots       & 195                               & 195                              \\ \bottomrule
\end{tabular}
\end{table}

\subsubsection{Comparison between \textbf{CS-SHRED} and \textbf{SHRED} for the Rotating Turbulent Flow}

\begin{figure}
     \centering
     \begin{subfigure}[b]{0.47\textwidth}
         \centering
         \includegraphics[width=\textwidth]{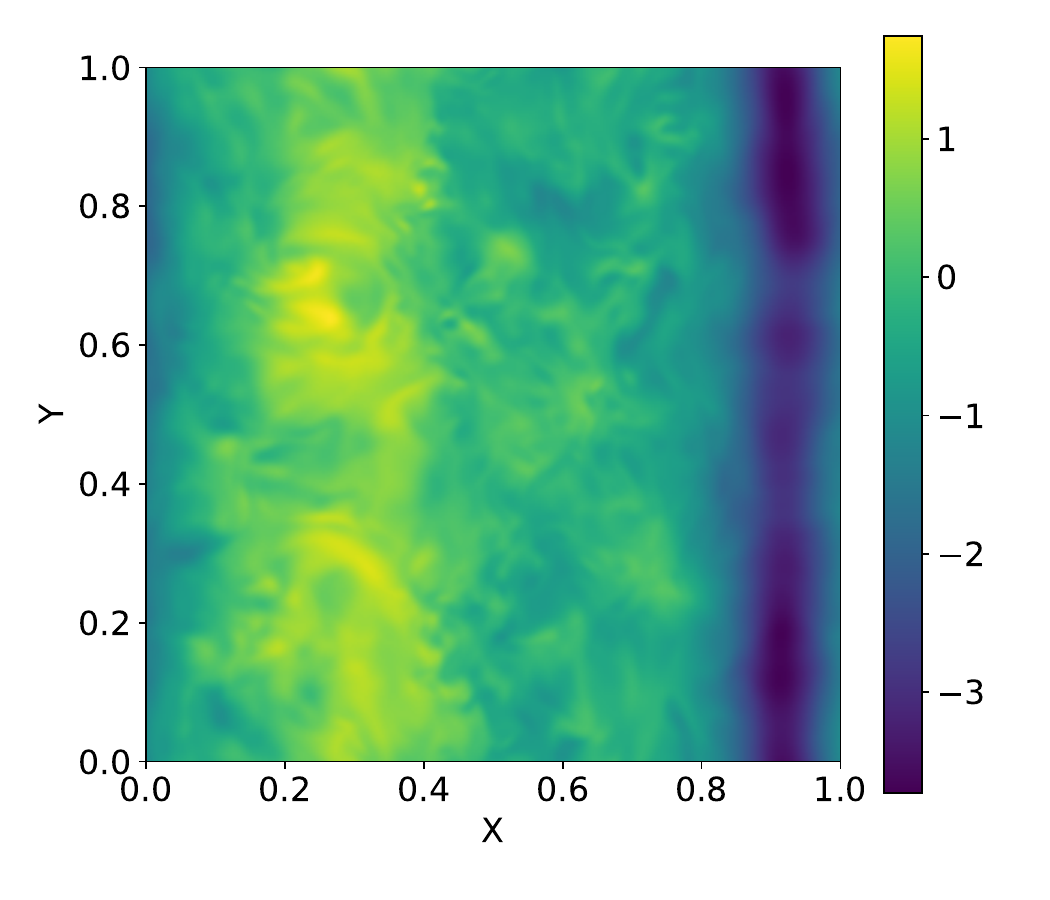}
         \caption{Original distribution}
         \label{fig:rot_ori}
     \end{subfigure}
     \hfill
     \begin{subfigure}[b]{0.47\textwidth}
         \centering
         \includegraphics[width=\textwidth]{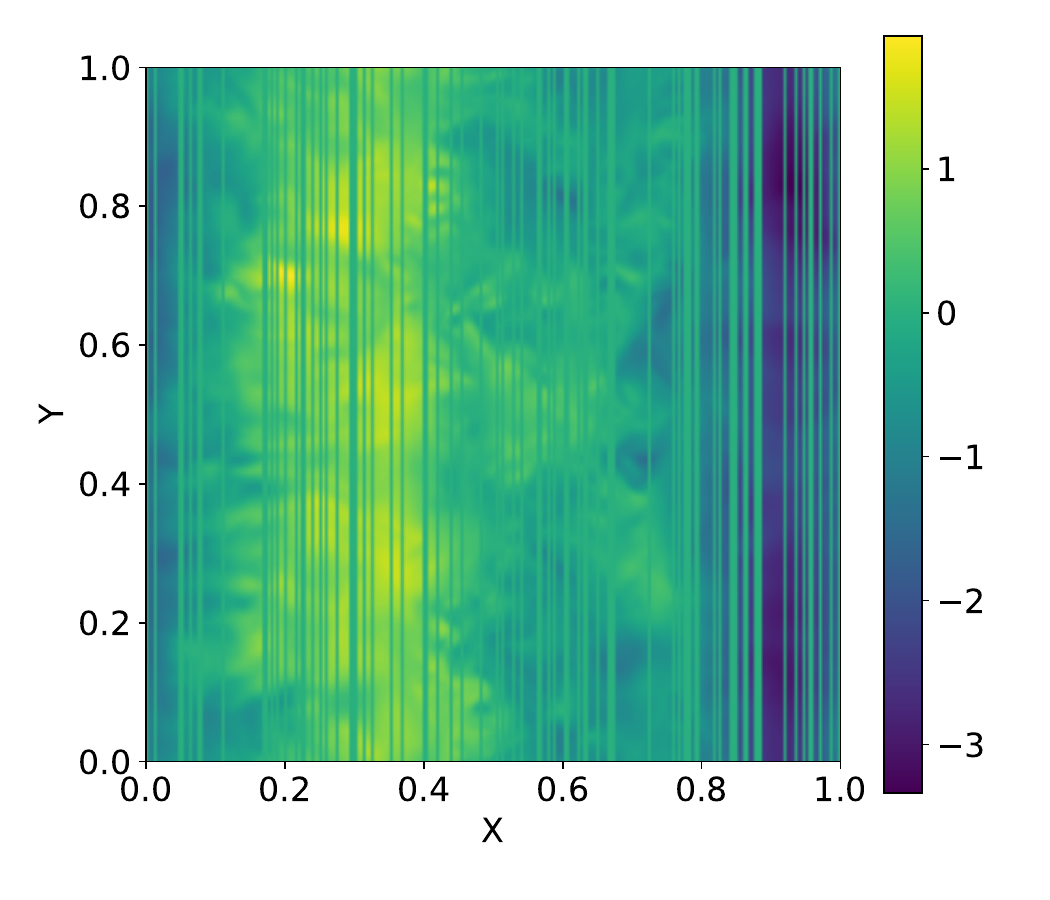}
         \caption{Subsampled distribution}
         \label{fig:rot_sub}
     \end{subfigure}
     \hfill
     \begin{subfigure}[b]{.95\textwidth}
         \centering
         \includegraphics[width=\textwidth]{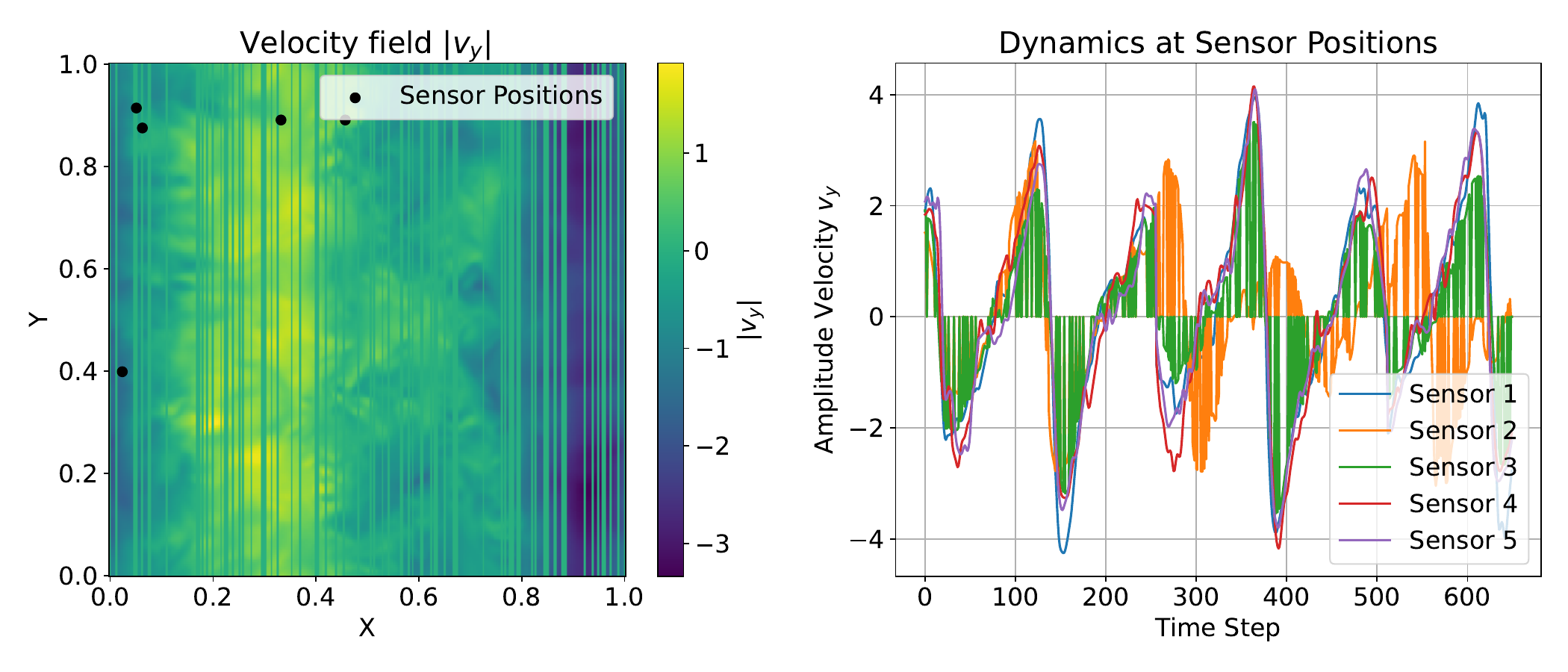}
         \caption{Temporal dynamics over 650 snapshots}
         \label{fig:rot_din}
     \end{subfigure}
     \hfill
     \begin{subfigure}[b]{0.47\textwidth}
         \centering
         \includegraphics[width=\textwidth]{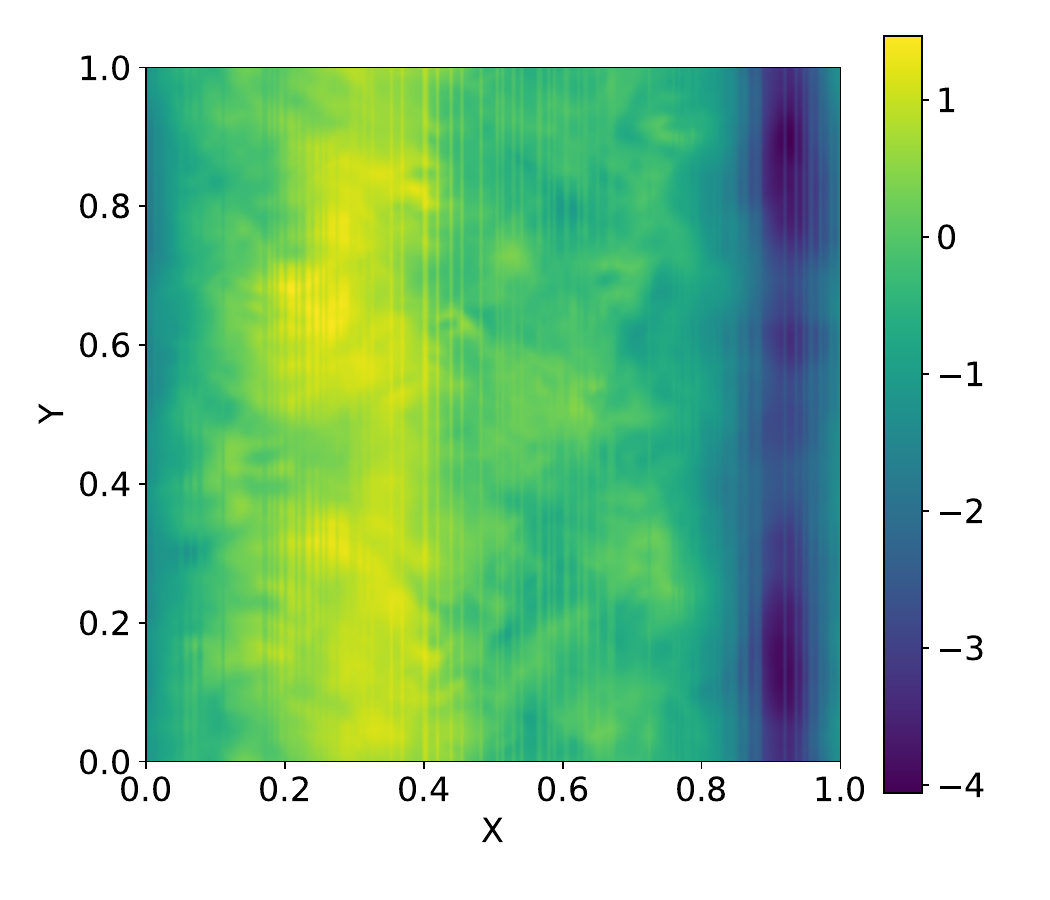}
         \caption{\textbf{SHRED}}
         \label{fig:rot_sh}
     \end{subfigure}
      \hfill
     \begin{subfigure}[b]{0.47\textwidth}
         \centering
         \includegraphics[width=\textwidth]{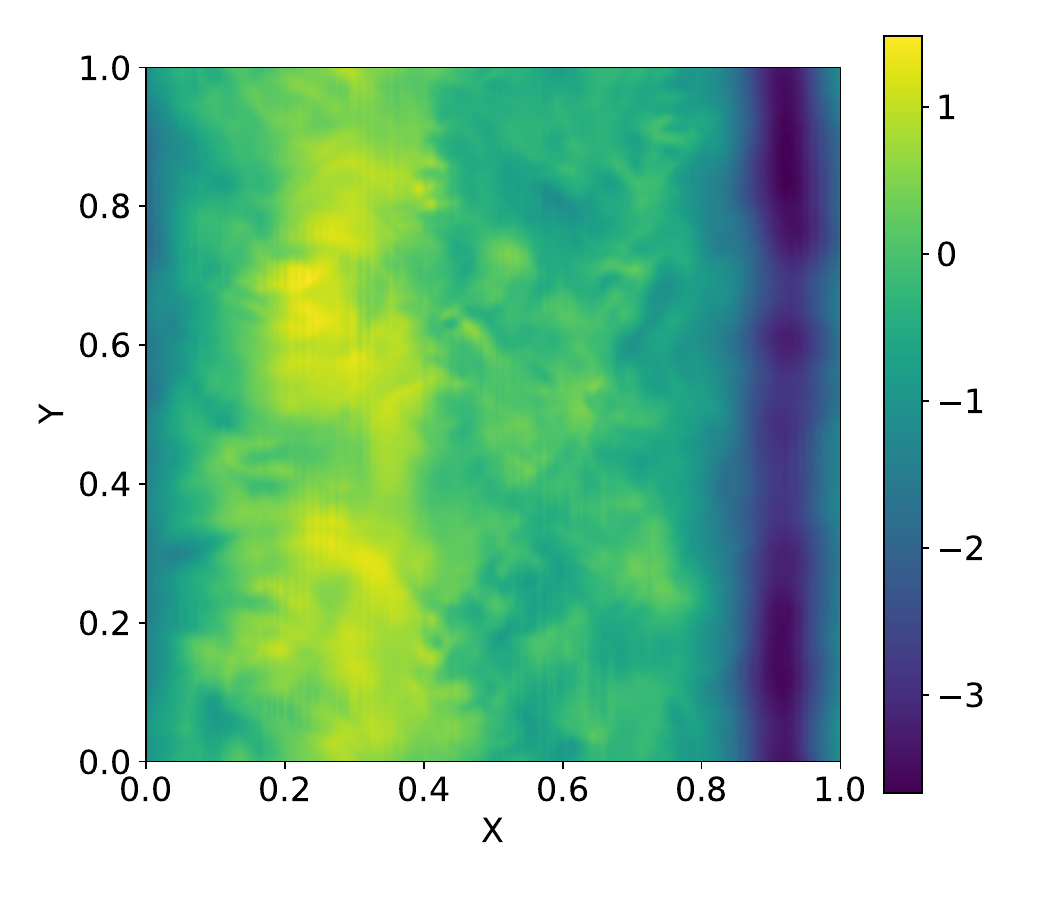}
         \caption{\textbf{CS-SHRED}}
         \label{fig:rot_cs}
     \end{subfigure}
        \caption{Comparison between \textbf{SHRED} and \textbf{CS-SHRED} for the Rotating Turbulent Flow data: (a) shows the original data of the last snapshot, (b) shows the respective subsampled data, (c) shows the temporal dynamics captured by a sensor fixed at a randomly chosen spatial position. The resultant reconstructions are shown at (d) for \textbf{SHRED} and (e) for \textbf{CS-SHRED}.}
        \label{fig:rot_ori_sub}
\end{figure}

To simulate practical measurement constraints, we randomly subsampled the dataset by removing 30\% of the columns in 30\% of the snapshots from a total of 650 snapshots with spatial dimensions $256 \times 256$. Figures~\ref{fig:rot_ori} and \ref{fig:rot_sub} show the spatial distribution of a velocity field slice before and after subsampling, respectively.

Figure~\ref{fig:rot_din} illustrates the temporal evolution at a fixed spatial point, emphasizing the challenges in reconstructing the turbulent flow from sparse and incomplete measurements.

Table~\ref{tab:metrics_rot_flow} summarizes the quantitative metrics used to compare the performance of \textbf{CS-SHRED} and \textbf{SHRED}. Metrics include Normalized Error, LPIPS, SSIM, and PSNR. For the final snapshot, \textbf{CS-SHRED} achieves markedly lower Normalized Error and LPIPS, as well as higher SSIM and PSNR, compared to \textbf{SHRED}. Over the entire time sequence, \textbf{CS-SHRED} exhibits slightly higher mean SSIM and PSNR, indicating superior overall performance.

\begin{table}[t]
\centering
\caption{Quantitative metrics comparing \textbf{CS-SHRED} and \textbf{SHRED} for the rotating turbulent flow, with ideal ranges}
\label{tab:metrics_rot_flow}
\begin{tabular}{@{}lccc@{}}
\toprule
\textbf{Metric}                        & \textbf{CS-SHRED} & \textbf{SHRED} & \textbf{Ideal Value} \\ \midrule
\textbf{Metrics (Lower is Better)}     &                   &                &                      \\
\quad Normalized Error                 & 0.0723            & 0.1559         & 0                    \\
\quad LPIPS                            & 0.0663            & 0.3720         & 0                    \\ \midrule
\textbf{Metrics (Higher is Better)}    &                   &                &                      \\
\quad Mean SSIM                        & 0.6437            & 0.6223         & 1                    \\
\quad Mean PSNR (dB)                   & 20.37             & 20.16          & Higher is better     \\
\quad SSIM (Last Snapshot)             & 0.9090            & 0.6918         & 1                    \\
\quad PSNR (Last Snapshot) (dB)        & 25.95             & 19.28          & Higher is better     \\ \bottomrule
\end{tabular}
\end{table}

Figures~\ref{fig:rot_cs} and \ref{fig:rot_sh} show the reconstructed velocity field for the final snapshot as obtained by \textbf{CS-SHRED} and \textbf{SHRED}, respectively. While \textbf{CS-SHRED} achieves a more detailed reconstruction at the final time step (supported by its high SSIM and low LPIPS), it also demonstrates higher fidelity over the entire sequence, as indicated by its mean SSIM and PSNR values.

\section{Final Considerations}

The proposed \textbf{CS-SHRED} architecture offers a robust and effective solution for reconstructing spatiotemporal dynamics from incomplete, compressed, or corrupted data. By integrating CS into the Shallow Recurrent Decoder (\textbf{SHRED}), our approach leverages a batch-based forward framework with $\ell_1$ minimization, making it particularly well-suited to scenarios with sparse sensor deployments, noisy measurements, and incomplete acquisitions.

A principal innovation of \textbf{CS-SHRED} is twofold. First, the incorporation of CS techniques into the \textbf{SHRED} architecture enables efficient recovery from undersampled data by exploiting the inherent sparsity of the signal. Second, the adaptive loss function, which dynamically combines MSE and MAE terms with a piecewise Signal-to-Noise Ratio (SNR) regularization, plays a critical role in suppressing noise and outliers in low-SNR regions while preserving essential small-scale structures in high-SNR regions. Although not implemented in this study, the integration of frequency-domain constraints (e.g., bandpass filtering) is a viable enhancement that could further emphasize critical spectral components of the underlying phenomena, especially in complex environmental and fluid dynamic systems.

Our extensive validation across a variety of applications---including viscoelastic fluid flows, maximum specific humidity fields, sea surface temperature distributions, and rotating turbulent flows---demonstrates that \textbf{CS-SHRED} consistently outperforms the traditional \textbf{SHRED} approach. This superiority is reflected in higher SSIM and PSNR values, lower normalized errors, and improved LPIPS scores, underscoring the architecture's enhanced reconstruction fidelity and robustness.

The successful integration of compressive recovery with recurrent modeling in \textbf{CS-SHRED} opens new avenues for accurate and reliable data reconstruction in environments with limited or corrupted measurements. Future work may further refine the approach by incorporating additional frequency-domain constraints and extending its application to even more challenging spatio-temporal datasets, thereby broadening its impact in environmental, climatic, and scientific data analyses.

\section*{Author Contributions}

\textbf{Romulo B.\ da Silva}: Conceptualization, Methodology, Software, Formal Analysis, Data Curation, Validation, Visualization, Writing—Original Draft, Writing—Review \& Editing.

\textbf{Cássio M.\ Oishi}: Supervision, Project Administration, Resources, Funding Acquisition, Writing—Review \& Editing.

\textbf{Diego Passos}: Visualization (figure organization and standardization), Writing—Review \& Editing.

\textbf{J.\ Nathan Kutz}: Senior Review, Advisory Support, Writing—Review \& Editing.

\vspace{0.5em}
\noindent
All authors have read and approved the final manuscript.

\section*{Acknowledgments}
    
The authors would like to thank the National Council for Scientific and Technological Development (CNPq) for financial support through research grants, and LSNIA (https://lsnia-unesp.github.io/) for the computational support. The CMO acknowledges support from the Sao Paulo Research Foundation (FAPESP).
The work of JNK was supported in part by the US National Science Foundation (NSF) AI Institute for Dynamical Systems (dynamicsai.org), grant 2112085. JNK further acknowledges support from the Air Force Office of Scientific Research  (FA9550-24-1-0141).

\section*{Code and Data Availability}
The source code used in this study is publicly available at \url{https://github.com/romulobrito/cs-shred}.

\newpage
\bibliographystyle{unsrt}  
\bibliography{references}

\end{document}